%% file: anonymous-submission-latex-2026.tex
\documentclass[letterpaper]{article} %
\usepackage{aaai2026}  %
\usepackage{times}  %
\usepackage{helvet}  %
\usepackage{courier}  %
\usepackage[hyphens]{url}  %
\usepackage{graphicx} %
\def\UrlFont{\rm}  %
\usepackage{natbib}  %
\usepackage{caption} %
\usepackage{algorithm}
\usepackage{algorithmic}
\usepackage{subcaption}
\usepackage{amsmath}
\usepackage{amsfonts}
\usepackage{booktabs}
\usepackage{multirow}
\usepackage{wasysym}
\usepackage{newfloat}
\usepackage{listings}
\DeclareCaptionStyle{ruled}{labelfont=normalfont,labelsep=colon,strut=off} %
\floatstyle{ruled}
\newfloat{listing}{tb}{lst}{}
\floatname{listing}{Listing}
\nocopyright 
\title{SimROD: A Simple Baseline for Raw Object Detection with Global and Local Enhancements}
\author{
    Haiyang Xie\textsuperscript{\rm 1,2},
    Xi Shen\textsuperscript{\rm 2},
    Shihua Huang\textsuperscript{\rm 2},
    Qirui Wang\textsuperscript{\rm 1,2},
    Zheng Wang\textsuperscript{\rm 1}\thanks{Corresponding author}
}
\affiliations{
    \textsuperscript{\rm 1}National Engineering Research Center for Multimedia Software, School of Computer Science, Wuhan University\\
    \textsuperscript{\rm 2}Intellindust AI Lab\\
    \texttt{\small \{whuocean,wangzwhu\}@whu.edu.cn, \{shenxiluc,shihuahuang95\}@gmail.com, wangqirui1937@163.com}
}

\usepackage{bibentry}

\begin{document}

\maketitle
\thispagestyle{plain}
\pagestyle{plain}

\begin{abstract}
Most visual models are designed for sRGB images, yet RAW data offers significant advantages for object detection by preserving sensor information before ISP processing. This enables improved detection accuracy and more efficient hardware designs by bypassing the ISP. However, RAW object detection is challenging due to limited training data, unbalanced pixel distributions, and sensor noise. To address this, we propose SimROD, a lightweight and effective approach for RAW object detection. We introduce a Global Gamma Enhancement (GGE) module, which applies a learnable global gamma transformation with only four parameters, improving feature representation while keeping the model efficient. Additionally, we leverage the green channel's richer signal to enhance local details, aligning with the human eye’s sensitivity and Bayer filter design. Extensive experiments on multiple RAW object detection datasets and detectors demonstrate that SimROD outperforms state-of-the-art methods like RAW-Adapter and DIAP while maintaining efficiency. Our work highlights the potential of RAW data for real-world object detection. 
\end{abstract}
\begin{links}
    \link{Project page}{https://ocean146.github.io/SimROD2025/}
    \link{Extended version}{https://arxiv.org/abs/2503.07101}
\end{links}

\section{Introduction}
\label{sec:intro}

\begin{figure}[!t]
    \label{fig:teaser}
    \centering
    \begin{minipage}{\linewidth}
        \begin{subfigure}{0.95\textwidth}
            \centering
            \includegraphics[width=1.0\linewidth]{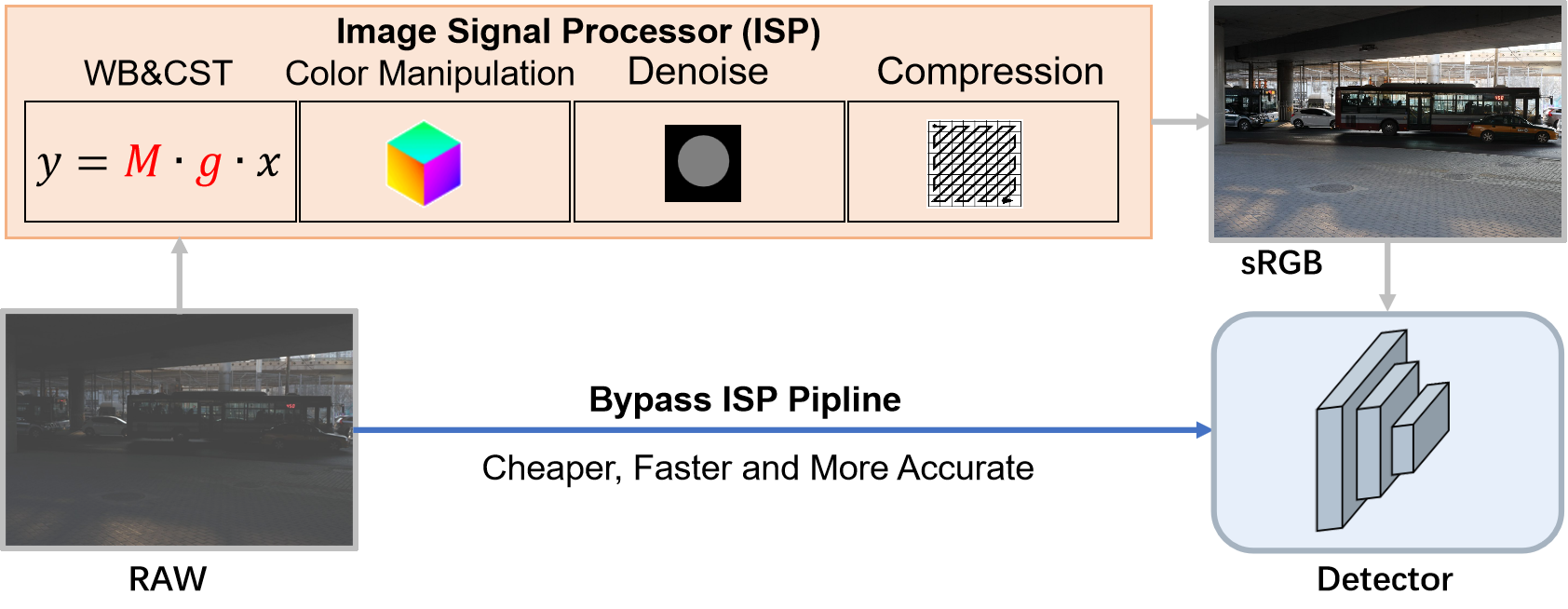}
            \caption{\textbf{Raw object detection enables bypassing the ISP.}}
            \label{fig_teaser_a}
        \end{subfigure}
        \begin{subfigure}{0.95\textwidth}
            \centering
            \includegraphics[width=1.0\linewidth]{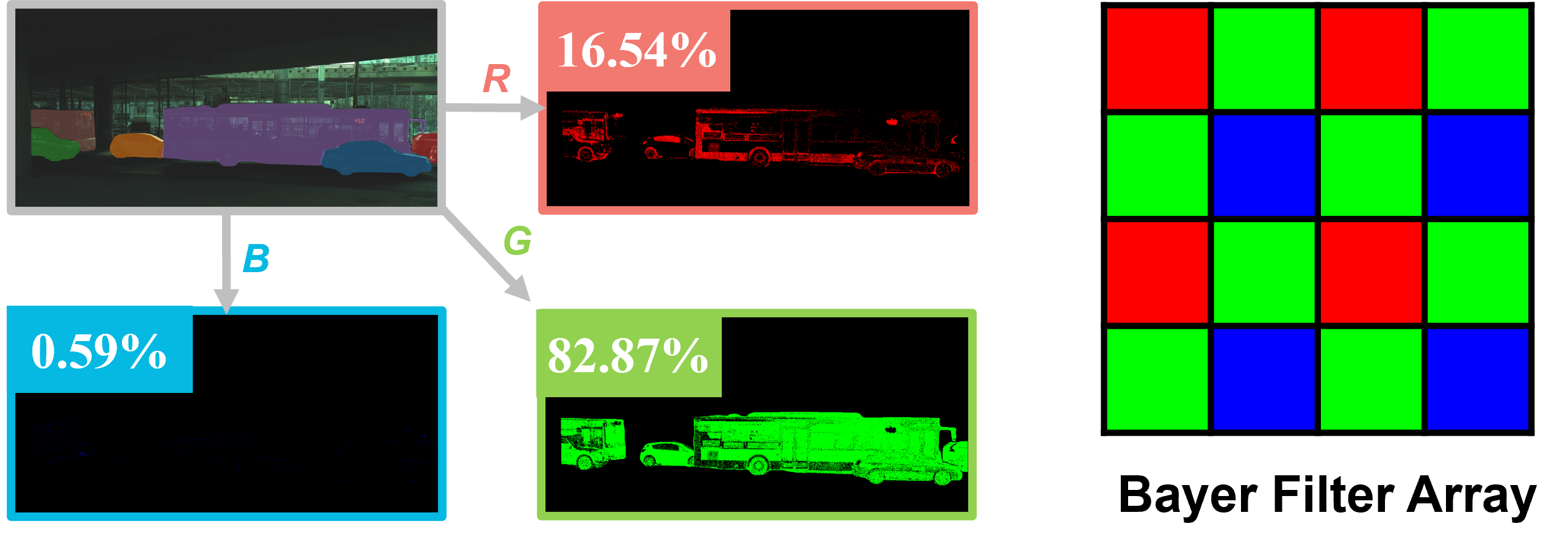}
            \caption{\textbf{Green channels contain more informative signals.}}
            \label{fig_teaser_b}
        \end{subfigure}
    \end{minipage}
    \caption{\textit{Top:} \textbf{Advantages of RAW Data for Object Detection.} Using RAW data eliminates the need for an ISP, reducing system complexity, latency, and cost—crucial for lightweight, real-time applications (Figure~\ref{fig_teaser_a}). \textit{Bottom:} \textbf{Key Insights in SimROD.} The green channel in RAW data carries more detailed information. The percentages indicate the proportion of color pixels with the highest intensity in the RGB channels—higher values mean richer details and lower noise in challenging lighting conditions (Figure~\ref{fig_teaser_b}).}
\end{figure}

Accurate object detection is crucial for autonomous driving, especially under challenging lighting and weather conditions. Traditional methods that rely on sRGB images often lose important details during processing. In contrast, RAW sensor data captures the unprocessed, richer signal from the sensor, preserving more details and a wider dynamic range~\cite{xu2023toward,raw_adapter,morawski2022genisp,chen2024guiding}. Moreover, as shown in Figure ~\ref{fig_teaser_a}, by directly using RAW data, there's no need for an ISP module, which can reduce system complexity, lower latency, and cut costs—key benefits for lightweight, real-time applications.

However, working with RAW data introduces several challenges, including limited training samples, unbalanced pixel distributions, and sensor noise. Current approaches typically rely on complex frameworks that integrate end-to-end optimization of ISP stages with object detection models. These methods explicitly designs learnable ISP stages to transform RAW data \cite{xu2023toward, morawski2022genisp, mosleh2020hardware, yu2021reconfigisp}. 
While these methods demonstrate promising results, they tend to be computationally expensive and introduce unnecessary design complexities.
Additionally, modern cameras emphasize the green channel in their Bayer filter design~\cite{zou2023rawhdr,bayer_uspat} as the human eye is highly sensitive to green light in both bright and low-light conditions~\cite{wald1945human}. However, most existing methods treat RGB channels equally, overlooking the green channel’s unique advantages in RAW data.

In this work, we present SimROD, a simple yet effective approach that enhances RAW object detection performance while maintaining model simplicity. Our approach is based on two key insights: (1) learning an adapted global transformation might not be complicated but crucial for fine-grained tasks \cite{xu2023toward, buckler2017reconfiguring}, and (2) the superior informativeness of the green channel (Figure~\ref{fig_teaser_b}) in the RGGB Bayer pattern. By leveraging these insights, we introduce an efficient Global Gamma Enhancement (GGE) with only four learnable parameters, significantly reducing model complexity while achieving comparable performance to more complex methods. We also propose a Green-Guided Local Enhancement (GGLE) module that uses the green channel to refine local image details, further boosting detection accuracy.

Through extensive experiments, we demonstrate that SimROD outperforms existing methods, such as RAW-Adapter~\cite{raw_adapter} and DIAP \cite{xu2023toward}, on several standard RAW object detection benchmarks, including ROD \cite{xu2023toward}, LOD \cite{hong2021crafting}, and Pascal-Raw~\cite{omid2014pascalraw}. For example, on the benchmark of Pascal-Raw \cite{omid2014pascalraw}, following the setup of RAW-Adapter~\cite{raw_adapter}, we achieve consistent performance improvement across different object detectors and different setups of Pascal-Raw \cite{omid2014pascalraw}.
Furthermore, we create a strong baseline for DIAP~\cite{xu2023toward} by leveraging weights pre-trained on MS COCO~\cite{lin2014microsoft}, which raises its performance from 24.0\% mAP to 30.7\% mAP on the ROD dataset~\cite{xu2023toward}. SimROD achieves notable improvements even relative to this strong baseline.

To summarize, the main contributions of this work are as follows:

\begin{itemize}

\item We introduce SimROD, a simple yet effective approach for RAW object detection that combines global-to-local enhancements.

\item Inspired by human visual system sensitivity and camera design, we confirm the informativeness of the green channel and develop a Green-Guided Local Enhancement module to refine local details and improve detection performance.

\item Despite its simplicity, SimROD achieves state-of-the-art performance on ROD \cite{xu2023toward}, LOD \cite{hong2021crafting}, and Pascal-Raw \cite{omid2014pascalraw}, surpassing prior methods like RAW-Adapter~\cite{raw_adapter} and DIAP \cite{xu2023toward}. \end{itemize}

\section{Motivation}
The human visual system exhibits a pronounced sensitivity to green light wavelengths in both bright and dim light conditions, as evidenced by~\cite{wald1945human}. Due to this characteristic of human vision, cameras prioritize green channels in their Bayer filter design~\cite{bayer_uspat,zou2023rawhdr}. Motivated by this biological and technical precedent, we explored the effectiveness of green channel for object detection by analyzing channel sensitivity and signal-to-noise ratio (SNR) for individual channels.

\begin{figure}[!t]
    \centering
    \includegraphics[width=\linewidth]{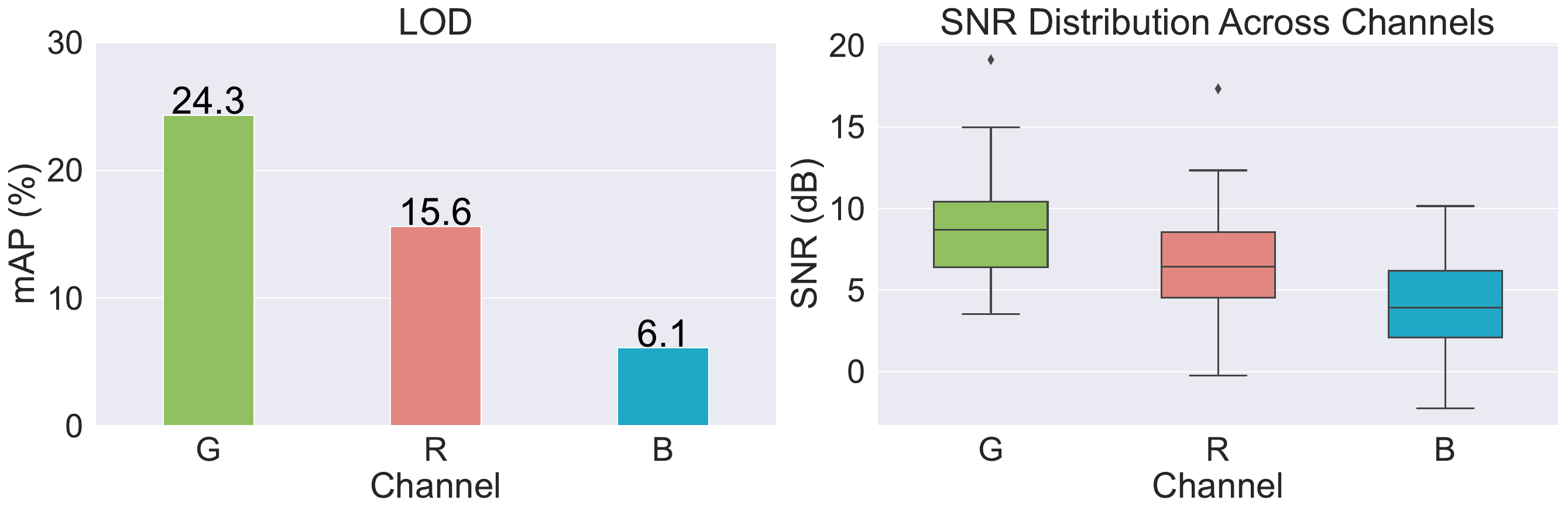}
    \caption{\textit{Left:} We evaluate RAW object detection on the LOD dataset~\cite{hong2021crafting} using individual color channels—green (G), red (R), and blue (B)—with the state-of-the-art DIAP method~\cite{xu2023toward}. The results highlight the superior performance of G. \textit{Right:} G has a significantly higher SNR than R and B, suggesting it may be more resistant to noise in extreme lighting conditions, potentially improving robustness.}
    \label{fig:motivation}
\end{figure}

\begin{itemize}
    \item \textit{Channel Sensitivity Analysis.} We utilized DIAP~\cite{xu2023toward} on the LOD~\cite{hong2021crafting} dataset to independently assess the detection performance of each channel (green, red, and blue). 
    As depicted in Figure~\ref{fig:motivation} \textit{left}, the green channel achieved the highest detection accuracy, surpassing the red and blue channels by substantial margins (approximately 10 and 20 AP points, respectively), underscoring its superior informativeness for object detection in RAW data.

    \item \textit{Signal-to-Noise Ratio (SNR) Analysis.}
    As presented in Figure~\ref{fig:motivation} \textit{right}, the green channel consistently exhibits a higher SNR compared to the red and blue channels, suggesting it is less susceptible to noise, even under challenging lighting conditions. This robustness reinforces the efficacy of leveraging green channel guidance to improve object detection accuracy in extreme environments.
\end{itemize}
These findings underscore the green channel’s potential to improve detection reliability, particularly in complex environments. Inspired by its demonstrated informativeness, we investigate a simple yet effective approach to fully leverage the green channel’s strengths, thereby enhancing model performance.

\section{Method}

\begin{figure*}[!htbp]
    \centering
    \includegraphics[width=1.0\linewidth]{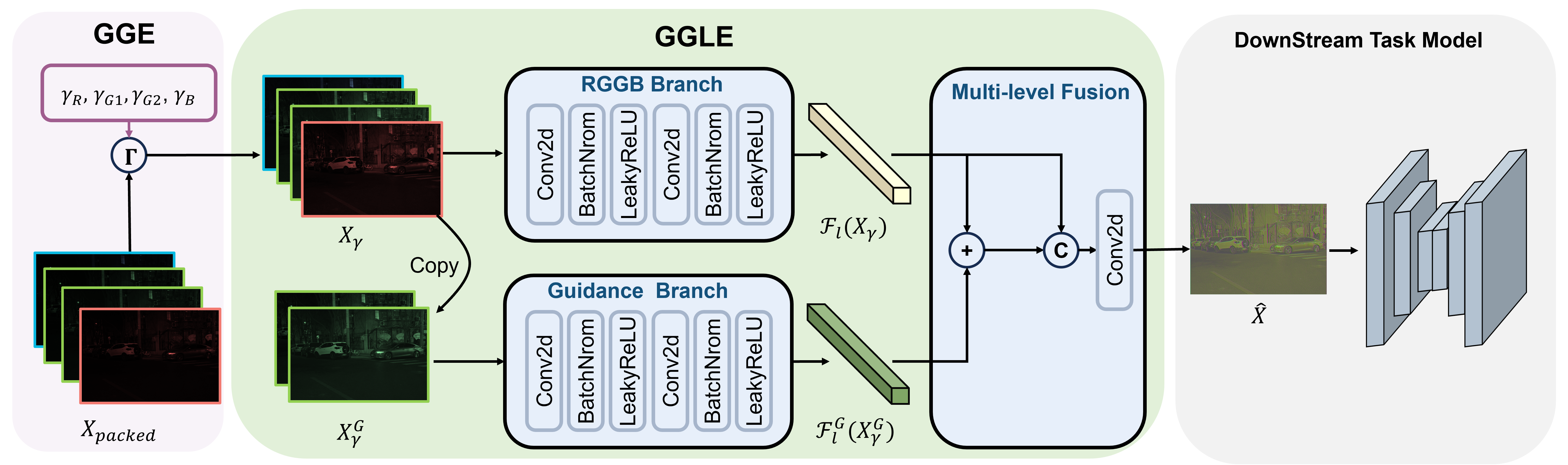}
    \caption{\textbf{The overview of our proposed SimROD.} Our SimROD takes a packed RAW image as input and first learns a global gamma transformation through the Global Gamma Enhancement (GGE) module. The transformed data is then processed by Green-Guided Local Enhancement (GGLE) to enhance local details. }
    \label{fig_framework2}
\end{figure*}

 The proposed method's overall framework is illustrated in Figure \ref{fig_framework2}. A RAW image is an unprocessed digital photograph that retains all the data from a camera's sensor, including the RGGB color pattern. Given a RAW image $( X_{\text{RAW}} \in \mathbb{R}^{2H \times 2W} $, we repack and convert it into a four-channel image \( X_{\text{packed}} \in \mathbb{R}^{H \times W \times 4} \), where the last dimension represents the color channels in the RGGB pattern.

We first adjust the global pixel distribution of \( X_{\text{packed}} \) using the proposed \textit{Global Gamma Enhancement} (GGE) module, which learns gamma transformation for each channel, resulting in \( X_{\gamma} \in \mathbb{R}^{H \times W \times 4} \). This \( X_{\gamma} \) is then passed into the proposed \textit{Green-Guided Local Enhancement} (GGLE) module for local region enhancement, producing an enhanced image \(\hat{X} \in \mathbb{R}^{H \times W \times 3} \). Finally, \( \hat{X} \) is fed into a downstream task model.

 This section is organized as follows: in Section~\ref{sec:gga}, we provide a detailed description of the proposed the \textit{Global Gamma Enhancement} (GGE) module; in Section~\ref{sec:ggle}, we detail the module \textit{Green-Guided Local Enhancement} (GGLE); implementation details are provided in Section~\ref{sec_imp_details}.

\subsection{Global Gamma Enhancement}
\label{sec:gga}

In visual perception tasks involving raw sensor data, pixel values are typically concentrated in the low range, which makes it challenging for deep neural networks to learn effectively and extract useful features~\cite{xu2023toward}. Therefore, dynamic range adjustment is an essential step in image signal processing pipelines to prepare raw data for object detection~\cite{xu2023toward, buckler2017reconfiguring}.

To address this issue, we propose a simple yet effective module named \textit{Global Gamma Enhancement} (GGE). 
For a four-channel packed raw image $X_{\text{packed}} \in \mathbb{R}^{H \times W \times 4}$, with pixel values normalized to the range $[0, 1]$, we assign a learnable gamma parameter for each channel. 
For the \textit{i}-th channel, the gamma transformation is defined as:
\begin{equation} \label{equ_gamma} X_{\gamma}^{i} = \Gamma(X_{\text{packed}}^i; \gamma_i), \quad i \in {R, G_1, G_2, B} \end{equation}
where $\gamma_i$ is simply a learnable parameter. Each channel is adjusted through a gamma transformation $\Gamma$
scaled to the range \([0, 255]\), calculated as:
\begin{equation}
\label{equ_gamma_trans}
\Gamma(X_{\text{packed}}^i; \gamma_i) = 255 \cdot \left( X_{\text{packed}}^i \right)^{\gamma_i}
\end{equation}
\paragraph{Discussion} In contrast to the recent method introduced by \cite{xu2023toward}, which predicts parameters for image-level RAW data adjustment, our proposed GGE contains only \textit{four} parameters, thus the network only consists of a minimal number of parameters. This results in greater computational efficiency while achieving comparable or superior performance to \cite{xu2023toward} (Section~\ref{sec_analysis}). Notably, we observed that the gamma parameters predicted by the image-level adjustment \cite{xu2023toward} module remained largely unchanged, even when a completely random noise image was used as input.

\subsection{Green-Guided Local Enhancement} 
\label{sec:ggle}

The \textit{Green-Guided Local Enhancement} (GGLE) module is designed to improve feature representations by exploiting the high-frequency details prevalent in the green channels of RAW data, formatted in the RGGB Bayer pattern. Specifically, GGLE processes these green channels independently alongside the full RGGB data, generating an optimized output tailored for downstream tasks such as object detection.

As illustrated in Figure~\ref{fig_framework2}, GGLE consists of two primary branches. The first branch, the RGGB branch, processes the complete RGGB data \( X_\gamma \) using a convolutional neural network, \( \mathcal{F}_{l} \), which extracts spatial features from all channels to produce a feature map, \( \mathcal{F}_{l}(X_\gamma) \), representing the full image context. The second branch, the Guidance branch, specifically targets the two green channels, \( X_\gamma^{G_1} \) and \( X_\gamma^{G_2} \), which are concatenated and processed through another convolutional network, \( \mathcal{F}_{l}^{G} \), resulting in a green-focused feature map, \( \mathcal{F}_{l}^{G}(X_\gamma^G) \), where \( X_\gamma^G = [X_\gamma^{G_1}, X_\gamma^{G_2}] \).

The final output is generated by a multi-level fusion of \( \mathcal{F}_{l}^{G}(X_\gamma^G) \) and \( \mathcal{F}_{l}(X_\gamma) \), expressed as:
\begin{equation}
    \label{equ_decode}
    \hat{X} = \text{Conv}(\text{Concat}[\mathcal{F}_{l}(X_\gamma) + \mathcal{F}_{l}^{G}(X_\gamma^G), \mathcal{F}_{l}(X_\gamma)])
\end{equation}
Here, \(\text{Conv}\) represents a convolution, while \(\text{Concat}\) denotes feature concatenation. The resulting output, \(\hat{X}\), is a three-channel representation that integrates structural details from the green channels across the RGB spectrum, enhancing performance in tasks that require high spatial resolution, such as object detection and segmentation.

\subsection{Implementation Details}
\label{sec_imp_details}

In the \textit{Global Gamma Enhancement} (GGE) module, each \(\gamma_i\) is parameterized in a straightforward manner. For each \(\gamma_i\), we define a learnable parameter \(\alpha_i \in \mathbb{R}\), which is constrained to the range \((-1, 1)\) using the \textit{tanh} activation function. This output is then linearly scaled to lie within a predefined range \((\gamma_{\text{min}}, \gamma_{\text{max}})\), where \(\gamma_{\text{min}}\) and \(\gamma_{\text{max}}\) are hyperparameters. Following the settings in~\cite{xu2023toward}, we set \(\gamma_{\text{max}} = 1/7.0\) and \(\gamma_{\text{min}} = 1/10.5\).

For the \textit{Green-Guided Local Enhancement} (GGLE) module, both \(\mathcal{F}_l\) and \(\mathcal{F}_l^{G}\), used in the RGGB and Guidance branches respectively, employ a simple yet effective architecture consisting of convolutional layers, Batch Normalization~\cite{ioffe2015batch}, and LeakyReLU~\cite{nair2010rectified} activation functions. 
Collectively, GGE and GGLE comprise a total of just 0.003 million parameters, rendering these modules lightweight compared to previous approaches. In contrast, prior methods often require hundreds of times more parameters, yet our approach achieves superior performance.

\paragraph{Loss Function}
Our SimROD is an end-to-end framework, jointly optimizing the GGE and GGLE modules alongside the downstream model, thereby eliminating the need for additional loss functions tailored to these enhancement stages. Furthermore, we adopt the same loss functions as those employed in the original works~\cite{ge2021yolox,lin2017focal,xie2021segformer}.
For example, when using YoloX~\cite{ge2021yolox} as the detector, optimization relies solely on the standard detection loss, which includes classification and regression components. The total loss function is defined as:
\begin{equation}
\label{equ_det_loss}
    \mathcal{L}_{total} = \mathcal{L}_{cls} + \lambda \mathcal{L}_{reg}
\end{equation}
where \(\lambda\) is the default balancing factor of the detector between the classification loss \(L_{cls}\) and the regression loss \(L_{reg}\) (\(\lambda=3\) in YoloX~\cite{ge2021yolox}). This unified approach allows the enhancement modules to adapt naturally to the detection objectives, supporting end-to-end optimization of the entire framework.

\section{Experiments}

\begin{table*}[htbp]
  \centering
  \caption{\textbf{Results with YoloX-Tiny~\cite{ge2021yolox}, following DIAP's benchmark~\cite{xu2023toward}.} 
  Performance metrics (AP and AP$_{50}$) for YoloX-Tiny across different methods including IA~\cite{xu2023toward}, Raw-or-cook~\cite{ljungbergh2023raw}, GenISP~\cite{morawski2022genisp}, RAW-Adapter~\cite{raw_adapter} and DIAP~\cite{xu2023toward}.
  The best performance for each dataset is highlighted in bold. The table also includes the number of parameters (in millions). The Pascal-Raw~\cite{omid2014pascalraw} in this table is the normal light version. 
  $\dag$ indicates reproduced results.
  N/A means the model did not converge.}

  \begin{tabular}{cccccccc}
    \toprule
    \multirow{2}[4]{*}{Method} & \multicolumn{2}{c}{LOD} & \multicolumn{2}{c}{Pascal-Raw} & \multicolumn{2}{c}{ROD} & Add. \\
    \cmidrule{2-7} & AP & AP$_{50}$ & AP & AP$_{50}$ & AP & AP$_{50}$ & Params (M) \\
    \midrule
    Demosacing & 26.5 & 46.0 & 66.8 & 92.8 & 4.8 & 9.5 & 0.000 \\
    Gamma & 25.7 & 44.2 & 69.0 & 94.5 & 7.6 & 14.3 & 0.000 \\
    \cmidrule{1-8}
    IA & 25.1 & 43.9 & 68.9 & 94.2 & 30.6 & 53.3 & 0.176 \\
    Raw-or-Cook$\dag$ & 18.0 & 36.3 & 61.6 & 91.2 & - & - & 0.000 \\
    GenISP$\dag$ & 20.5 & 39.8 & 60.6 & 89.5 & - & - & 0.220 \\
    RAW-Adapter$\dag$ & 26.4 & 45.1 & 67.5 & 93.7 & N/A & N/A & 0.460 \\
    DIAP & 25.9 & 43.4 & 68.7 & 94.2 & 30.7 & 53.4 & 0.260 \\
    \cmidrule{1-8}
    \textbf{Our SimROD} & \textbf{26.7(+0.8)} & \textbf{46.3(+2.9)} & \textbf{69.7(+1.0)} & \textbf{95.1(+1.1)} & \textbf{33.1(+2.4)} & \textbf{57.6(+4.2)} & 0.003 (1\%) \\
    \bottomrule
  \end{tabular}
  \label{tab_sota_diap}
\end{table*}

\subsection{Datasets and Evaluation Metrics}
We evaluate and compare our method against existing methods on four benchmark datasets: Pascal-Raw~\cite{omid2014pascalraw}, LOD~\cite{hong2021crafting}, and ROD~\cite{xu2023toward} for object detection, and ADE20K-Raw~\cite{raw_adapter} for semantic segmentation.

\paragraph{Pascal-Raw~\cite{omid2014pascalraw}.} 
The Pascal-Raw dataset~\cite{omid2014pascalraw} comprises 4,259 RAW images captured under standard lighting conditions with a Nikon D3200 DSLR camera, covering three object classes: person, car, and bicycle. The dataset is divided into 2,129 images for training and 2,130 images for testing. For our experiments, we utilize the preprocessed RAW data provided by RAW-Adapter~\cite{raw_adapter}.

\paragraph{LOD~\cite{hong2021crafting}.} 
The LOD dataset~\cite{hong2021crafting} contains 2,230 RAW images captured with a Canon EOS 5D Mark IV camera in low-light conditions, covering eight object classes: bus, chair, TV monitor, bicycle, bottle, dining table, motorbike, and car. The dataset is split into 1,800 images for training and 430 images for testing.
For all our experiments, we use the preprocessed RAW data provided by RAW-Adapter~\cite{raw_adapter} to ensure a fair comparison with RAW-Adapter.

\paragraph{ROD~\cite{xu2023toward}.} 
The original ROD dataset~\cite{xu2023toward} contains 25,207 RAW images, including 10k daytime scenes and 14k nighttime scenes, across six common object categories. Compared to other datasets, ROD is a larger-scale dataset with a more diverse range of scenes, focusing on urban driving scenarios. Due to limitations in dataset access, we were unable to adhere to the ``standard" partitioning protocol defined by ROD~\cite{xu2023toward}. Despite repeated attempts to obtain the full dataset, only a subset of the training data has been publicly released. This subset consists of 16,089 RAW images, including 4,053 daytime scenes and 12,036 nighttime scenes, but it contains only five object classes instead of the six originally specified in~\cite{xu2023toward}. To ensure a fair evaluation, we randomly partitioned the publicly available subset into 80\% for training and 20\% for testing. This partition resulted in 3,245 daytime scenes and 9,626 nighttime scenes in the training set, with the remaining images reserved for testing. We strictly followed the official guidelines for data preprocessing and conducted multiple experimental trials to ensure the validity of the results. All references to the ROD dataset in this paper pertain specifically to this re-partitioned subset. 
We will make this subset publicly available to facilitate the re-implementation of our results.

\paragraph{ADE20K-Raw~\cite{raw_adapter}.} The ADE20K-RAW dataset~\cite{raw_adapter} is a RAW-format segmentation dataset derived from ADE20K~\cite{zhou2017scene}, consisting of 27,574 images synthesized using InvISP~\cite{xing2021invertible} by RAW-Adapter~\cite{raw_adapter}. This dataset includes three versions—normal, dark, and over-exposure—that simulate different lighting conditions. The dataset follows the same training and testing splits defined by ADE20K~\cite{zhou2017scene} for consistency.

\paragraph{Evaluation metrics.} 
For object detection, we report standard average precision at an IoU threshold of 0.5 (AP$_{50}$) and the average precision across IoU thresholds from 0.5 to 0.95 (AP). For semantic segmentation, we use mean Intersection over Union (mIoU), which measures the average overlap between predicted and ground truth masks across all classes.

\subsection{Training Details}
All experiments in this section follow the settings of DIAP~\cite{xu2023toward}, unless stated otherwise.

For detection tasks, we conduct experiments using two object detectors: YoloX~\cite{ge2021yolox} and RetinaNet~\cite{lin2017focal}, following the protocols of DIAP~\cite{xu2023toward} and RAW-Adapter~\cite{raw_adapter}. If not otherwise specified, all experiments were initialized with pre-trained weights.

For YoloX~\cite{ge2021yolox}, we use the official training strategies, including standard data augmentation techniques such as random horizontal flipping, scale jittering through resizing, and Mosaic augmentation~\cite{bochkovskiy2020yolov4}. Both training and testing data are resized to 640×640. The model is trained for 300 epochs, with five warmup epochs, using the SGD optimizer with a momentum of 0.9. We apply a cosine learning rate schedule and use a batch size of 12. The training process of Pascal-Raw~\cite{omid2014pascalraw} and LOD~\cite{hong2021crafting} takes 2 hours on three NVIDIA RTX 3090 GPUs. The training process of ROD~\cite{xu2023toward} takes 10 hours on three NVIDIA RTX 4090 GPUs. For initialization, we use COCO~\cite{lin2014microsoft} pre-trained weights, which improve the performance from 24.0\% AP to 30.7\% AP compared to the approach proposed in the original DIAP~\cite{xu2023toward} and construct a strong baseline.

For the RetinaNet~\cite{lin2017focal} detector, we use the MMDetection framework and its default data augmentation pipeline, which includes random cropping, random flipping, and multi-scale testing~\cite{raw_adapter}. We set the learning rate for the proposed SimROD to 3e-3. 

For the segmentation task, we followed the settings from RAW-Adapter~\cite{raw_adapter} and used Segformer~\cite{xie2021segformer} with the MIT-B5~\cite{xie2021segformer} backbone. We trained the model on four NVIDIA RTX 4090 GPUs, with learning rates adjusted empirically for the proposed method: 8e-5 for the normal-light dataset of Pascal-Raw~\cite{raw_adapter}, 7e-5 for the dark dataset of Pascal-Raw~\cite{raw_adapter}, and 9e-5 for the over-exposure dataset~\cite{raw_adapter}.

\subsection{Comparison with State-of-the-Art Methods}

\begin{table*}[htbp]
  \centering
  \caption{\textbf{Results with RetinaNet-R50~\cite{lin2017focal}, following RAW-Adapter's benchmark~\cite{raw_adapter}.} Performance metrics (AP and AP$_{50}$) for RetinaNet-R50 across different methods including Demosaicing~\cite{raw_adapter}, Default ISP~\cite{omid2014pascalraw}, Karaimer et Brown~\cite{karaimer2016software}, Lite-ISP~\cite{zhang2021learning}, InvISP~\cite{xing2021invertible} and Dirty-Pixel~\cite{diamond2021dirty}. The best performance for each dataset is highlighted in bold. The table also includes the number of parameters (in millions). The Pascal-Raw~\cite{omid2014pascalraw} in this table is the normal light version. $\dag$ indicates results reproduced using the official code. RetinaNet-R50 shows similar AP/AP$_{50}$ due to strong performance and low small-object ratios in LOD/Pascal-Raw (5\%/3\% vs. 42\% in COCO).}
    \begin{tabular}{cccccc}
    \toprule
    \multirow{2}[4]{*}{Method} & \multicolumn{2}{c}{LOD} & \multicolumn{2}{c}{Pascal-Raw} & Add. \\
    \cmidrule{2-5}
          & AP    & AP$_{50}$  & AP    & AP$_{50}$    & Params (M) \\
    \midrule
    Demosacing & 58.5  & 58.5  & 89.2  & 89.2      & 0.000  \\
    Default ISP & 58.4  & 58.4  & 89.6  & 89.6      & 0.000  \\
    \cmidrule{1-6}
    Karaimer et Brown & 54.4  & 54.4  & 89.4  & 89.4       & - \\
    Lite-ISP & -  & -  & 88.5     & 88.5     & -      \\
    InvISP & 56.9  & 56.9  & 87.6  & 87.6  & -      \\
    \cmidrule{1-6}
    Dirty-Pixel & 61.6  & 61.6  & 89.7  & 89.7       & 4.28 \\
    DIAP$\dag$ & 59.1 & 59.1 & 89.5 & 89.5  & 0.260 \\
    RAW-Adapter & 62.1  & 62.1  & 89.7  & 89.7       & 0.46  \\
    \cmidrule{1-6}
    \textbf{Our SimROD} & \textbf{63.4(+1.3)} & \textbf{63.4(+1.3)} & \textbf{90.1(+0.4)} & \textbf{90.1(+0.4)}     & 0.003 (0.7\%)  \\
    \bottomrule
    \end{tabular}
  \label{tab_sota_ra}
\end{table*}

\paragraph{RAW Object Detection on Pascal-Raw~\cite{omid2014pascalraw}, LOD~\cite{hong2021crafting}, and ROD~\cite{xu2023toward} Datasets.}
Table~\ref{tab_sota_diap} and Table~\ref{tab_sota_ra} demonstrate that our proposed SimROD consistently outperforms existing methods across all datasets while being highly efficient. To ensure a fair comparison with RAW-Adapter~\cite{raw_adapter} and DIAP~\cite{xu2023toward}, we strictly follow their official settings with YoloX-Tiny~\cite{ge2021yolox} and RetinaNet-R50~\cite{lin2017focal}. For YoloX-Tiny (Table~\ref{tab_sota_diap}), SimROD achieves notable AP gains of +2.4 on ROD, +0.8 on LOD, and +1.0 on Pascal-Raw, surpassing DIAP~\cite{xu2023toward} with just 0.003M additional parameters. For RetinaNet-R50 (Table~\ref{tab_sota_ra}), SimROD reaches an impressive 63.4\% AP$_{50}$ on LOD, outperforming RAW-Adapter~\cite{raw_adapter} and DIAP~\cite{xu2023toward}, while requiring only 0.7\% of RAW-Adapter’s additional parameters. These results highlight SimROD’s strong generalization, superior accuracy, and exceptional parameter efficiency, making it a highly effective solution for RAW object detection.

\paragraph{Semantic Segmentation on ADE20K-Raw~\cite{zhou2017scene,raw_adapter} with RAW Data.}
To further validate SimROD’s effectiveness, we evaluate it on semantic segmentation using ADE20K-Raw with Segformer~\cite{xie2021segformer}, following RAW-Adapter~\cite{raw_adapter}. As shown in Table~\ref{tab_ade20k}, SimROD achieves the best performance across normal and low-light conditions, while achieving competitive performance on over-exposure conditions, demonstrating strong generalization. Despite adding only 0.003M parameters, it remains highly efficient while delivering superior accuracy. These results reinforce our object detection findings, proving SimROD’s versatility in RAW data processing. Its potential extends beyond detection, making it applicable to broader vision tasks.

\begin{table}[t]
  \centering
  \setlength{\tabcolsep}{1mm}
  \caption{\textbf{Segmentation Performance on ADE20K-Raw~\cite{zhou2017scene,raw_adapter}.} Comparison of semantic segmentation (mIoU) results under normal, overexposure (\textit{Over-exp}), and low-light (\textit{Dark}) conditions using Segformer~\cite{xie2021segformer}. Bold numbers indicate the best results for each condition.}
    \begin{tabular}{cccccc}
    \toprule
    \multirow{2}{*}{Method} &  \multirow{2}{*}{Normal} &  \multirow{2}{*}{Over-exp} & \multirow{2}{*}{Dark} & Add. \\
    & & & & Params. (M)\\
    \midrule
    Demosacing & 47.47 & 45.69 & 37.55 & 0.00 \\
    Karaimer et Brown & 45.48 & 42.85 & 37.32 & 0.00 \\
    Lite-ISP & 43.22 & 42.01 & 5.52 & 0.00 \\
    InvISP & 47.82 & 44.30 & 4.03 & 0.00 \\
    SID & - & - & 37.06 & 0.00 \\
    DNF & - & - & 35.88 & 0.00 \\
    Dirty-Pixel & 47.86 & 46.50 & 38.02 & 4.28 \\
    RAW-Adapter & 47.95 & \textbf{46.62} & 38.75 & 0.30 \\
    \textbf{Our SimROD}  & \textbf{48.42} & 45.96 & \textbf{38.85} & 0.003 \\
    \bottomrule
    \end{tabular}
  \label{tab_ade20k}
\end{table}

\begin{table}[t]
  \centering
  \setlength{\tabcolsep}{1mm}
  \caption{
  \textbf{Ablation Study on the Impact of Different Components over Pascal-Raw~\cite{omid2014pascalraw} and LOD~\cite{hong2021crafting} datasets with YoloX-Tiny~\cite{ge2021yolox}.} The term `Guided Channel' refers to the use of a specific channel for local enhancement in GGLE, while GGLE without a guided channel indicates a configuration with only the RGGB branch. Bold numbers indicate the best performance.
  }
    \begin{tabular}{cccc|cc}
    \toprule
    Pretrained & \multicolumn{1}{c}{\multirow{2}{*}{GGE}} & \multicolumn{1}{c}{GGLE} & \multicolumn{1}{c}{GGLE} &  \multicolumn{1}{c}{LOD} & \multicolumn{1}{c}{Pascal-Raw} \\
     Weight&       &   (RGGB)    &  (Guidance)     & AP$_{50}$  & AP$_{50}$ \\
    \midrule 
    \Circle & \Circle & \Circle & \Circle & 27.7 & 85.0\\
    \CIRCLE & \Circle & \Circle & \Circle  & 44.6 & 93.0 \\
    \CIRCLE & \CIRCLE & \Circle & \Circle  & 45.0 & 94.3 \\
    \CIRCLE & \CIRCLE & \CIRCLE & \Circle & 45.1 & 94.7  \\
    \CIRCLE & \CIRCLE & \CIRCLE & R & 44.9  & 94.5  \\
    \CIRCLE & \CIRCLE & \CIRCLE & B & 44.2  & 94.2  \\
    \CIRCLE & \CIRCLE & \CIRCLE & RB & 44.5  & 94.3  \\
    \CIRCLE & \CIRCLE & \CIRCLE & RGGB & 46.2  & 94.5  \\
    \CIRCLE & \CIRCLE & \CIRCLE & GG & \textbf{46.3}  & \textbf{95.1}  \\
    \bottomrule
    \end{tabular}%
  \label{tab_res_pas_lod}
\end{table}

\begin{table}[!t]
  \centering
  \caption{\textbf{Comparison of DIAP~\cite{xu2023toward} and our GGE on ROD~\cite{xu2023toward}, Pascal-Raw~\cite{omid2014pascalraw}, and LOD~\cite{hong2021crafting} datasets with YoloX-Tiny~\cite{ge2021yolox}.} Results are shown in AP and AP$_{50}$ for each dataset, with additional parameters and GFLOPs also provided.}
  \begin{tabular}{lc|cc}
    \toprule
    Dataset & Metric & DIAP & Our GGE \\
    \midrule
    \multirow{2}{*}{ROD} & AP   &  30.7 &\bf 30.9 \\
                                             & AP$_{50}$  & 53.4 &\bf 53.5 \\
    \midrule
    \multirow{2}{*}{Pascal-Raw} & AP   &\bf 68.7 & 68.4 \\
                                                          & AP$_{50}$  & 94.2 & \bf 94.3 \\
    \midrule
    \multirow{2}{*}{LOD} & AP   & \bf 25.9 & 25.5 \\
                                                 & AP$_{50}$  & 43.4 & \bf 46.2 \\
    \midrule
    & Params (M) & 0.260 & 0.000 \\
    & GFLOPs     & 1.504 & 0.000 \\
    \bottomrule
  \end{tabular}%
  \label{tab_diap_gge}
\end{table}

\subsection{Analysis and Discussion}
\label{sec_analysis}

\paragraph{Ablation Study.}
Table~\ref{tab_res_pas_lod} presents an ablation study on the LOD~\cite{hong2021crafting}  and Pascal-Raw~\cite{omid2014pascalraw} datasets using YoloX-Tiny~\cite{ge2021yolox} to evaluate the impact of different enhancement modules. 
The best AP and AP$_{50}$ results are achieved when all four components are used, indicating the combined enhancements' positive effect on model performance. Moreover, the green channel guidance outperforms the red and blue guidance, indicating the unique value of the green channel. Notably, using the red and blue channel guidance even leads to the performance drop on LOD~\cite{hong2021crafting} which is low-light, noisy, and more challenging. We also evaluate RGGB Table~\ref{tab_res_pas_lod}. 
We find that that \textit{solely using the Green Channels (GG) alone outperforms the combination of all channels (RGGB)}. This is attributed to the significant noise in R and B of RGGB, which affects the learning of the model. The inferior performance of RB compared to R alone further corroborates this, suggesting that combining all channels as guidance does not yield better results.

\begin{table}[t]
\centering
\caption{\textbf{Effect of Green Channel Sampling Frequency.} Reducing the green channel sampling frequency to 0.5$\times$ leads to a significant performance drop, especially on the low-light LOD dataset, highlighting the importance of dense green channel information.}
\label{tab:sampling_f}
\begin{tabular}{lcccc}
\toprule
\multirow{2}{*}{G Sampling Frequency} & \multicolumn{2}{c}{LOD} & \multicolumn{2}{c}{Pascal-Raw} \\
\cmidrule(lr){2-3} \cmidrule(lr){4-5}
  & mAP & AP$_{50}$ & mAP & AP$_{50}$ \\
\midrule
0.5$\times$ (reduced) & 22.0 & 40.3 & 67.5 & 94.3 \\
1$\times$ (default)   & 24.3 & 42.0 & 67.9 & 94.2 \\
\bottomrule
\end{tabular}

\end{table}

\paragraph{GGE v.s. DIAP~\cite{xu2023toward}}
\label{sec_ana_rod_gge}
We compares DIAP~\cite{xu2023toward} and our GGE across ROD~\cite{xu2023toward}, Pascal-Raw~\cite{omid2014pascalraw}, and LOD~\cite{hong2021crafting} datasets in Table~\ref{tab_diap_gge}. Our GGE achieves comparable or slightly improved performance in certain metrics while reducing parameters and GFLOPs, reflecting a more efficient model design. Note that GGE only contains four learnable parameters.

\paragraph{Sampling Frequency of the Green Channel.}
We investigate how the green channel's sampling frequency affects RAW object detection performance (Table~\ref{tab:sampling_f}). In our experiments, we feed the green channel into DIAP~\cite{xu2023toward} and compare two strategies: the default 1× sampling based on the RGGB pattern versus a reduced 0.5× sampling (using only one green value). The results show that lowering the green channel frequency leads to a significant performance drop, especially on the low-light LOD dataset. This confirms that the denser spatial information provided by the green channel—due to its higher frequency in the Bayer pattern—is crucial for robust performance.

\begin{figure}[h]
    \centering
    \includegraphics[width=0.8\linewidth]
    {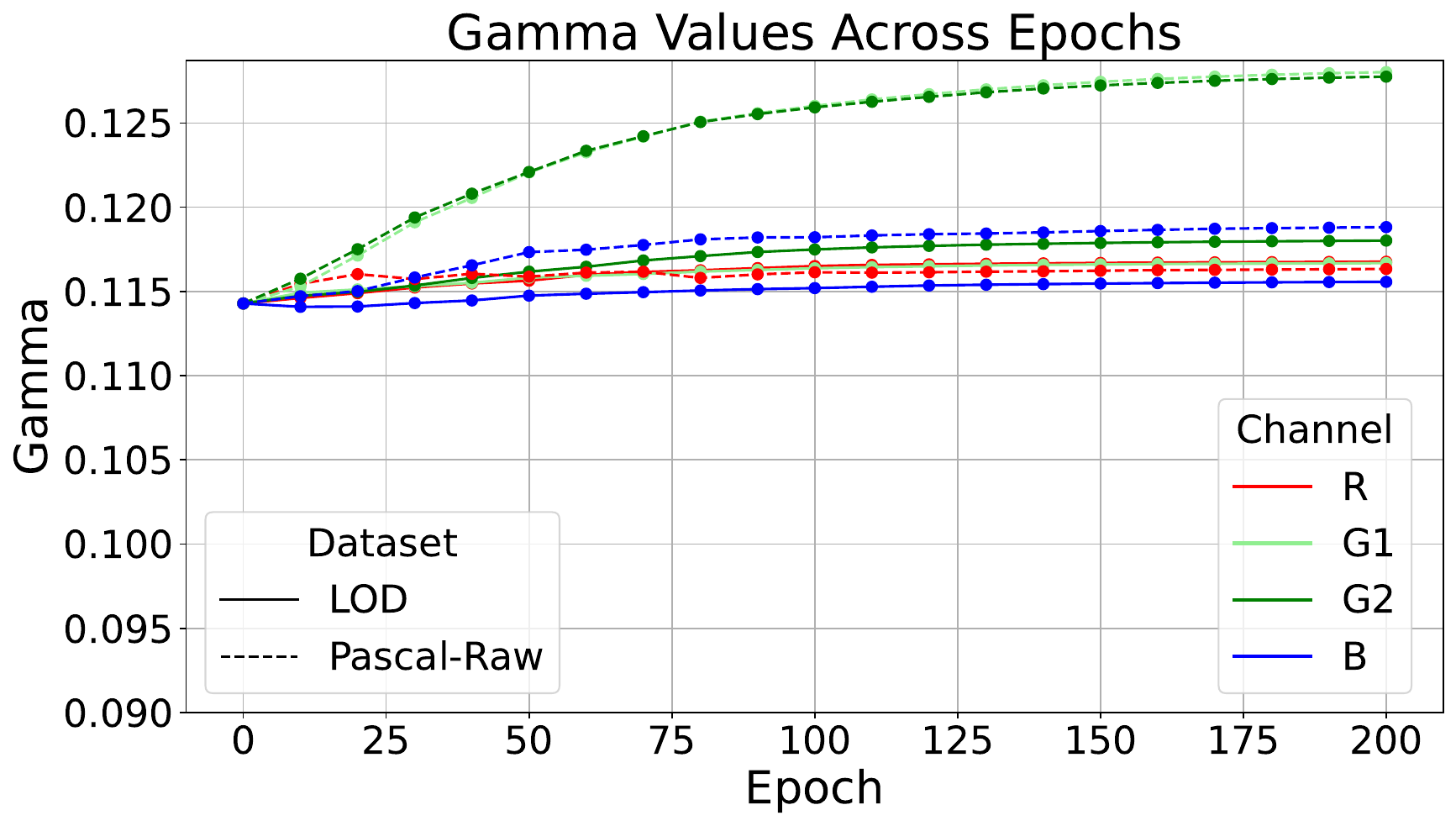}
    \caption{\textbf{$\gamma$ across epochs on LOD and Pascal-Raw.}}
    \label{fig:gamma_after_train}
\end{figure}

\paragraph{The Value of $\gamma$ in Training .} As shown in Fig.~\ref{fig:gamma_after_train}, the $\gamma$ values slightly increase during training leading to the resulting pixel values are slightly reduced. This is reasonable, as the initial value of 0.114 is lower than the commonly used default of 1/2.2 in standard gamma correction. 

\section{Conclusion}
In this work, we presented SimROD, a simple yet effective approach for enhancing object detection performance on RAW data. SimROD introduces a streamlined solution featuring the Global Gamma Enhancement (GGE) with only four learnable parameters, achieving competitive performance while maintaining low model complexity. Furthermore, our exploration revealed that the green channel holds more informative signals, leading to the development of the Green-Guided Local Enhancement (GGLE) module, which enhances local image details effectively. Extensive experiments across multiple RAW object detection datasets and detectors, as well as the RAW segmentation dataset, demonstrate the effectiveness of our SimROD.

\section{Related Work}
\label{sec:reltaed_work}

\paragraph{Object Detection on RAW Data}
Object detection has been an active area of research in computer vision~\cite{everingham2010pascal,lin2014microsoft,wu2020recent,Tian_2025_CVPR,Ma_2024_CVPR,wang2025leader,wang2025you}. RAW data serves as the input to the image signal processor (ISP) and has attracted significant attention due to its unique value.\cite{xu2023toward,karaimer2016software,Nishimura_Gerasimow_Sushma_Sutic_Wu_Michael_2018,Ramanath_Snyder_Yoo_Drew_2005}. 
Considerable efforts have been made to utilize the RAW data to improve detection performance and robustness. \cite{hong2021crafting} designed an auxiliary task for image reconstruction on synthetic datasets to improve detection performance. \cite{onzon2021neural} proposed estimating camera exposure parameters from previous frames to achieve well-exposed images. \cite{morawski2022genisp} designed an ISP pipeline called GenISP, which applies learnable white balance and color correction transformations to RAW data. \cite{yoshimura2023dynamicisp} decomposed control into the entire dataset, along with fine-tuning for individual images. They introduced a latent update style controller to manage the stages of a differentiable ISP. \cite{xu2023toward} developed an image-level adjustment module and a pixel-level adjustment module to learn transformations. \cite{chen2024guiding} proposed an activation function to extract features from RAW data effectively. \cite{raw_adapter} combined an input-level adapter with a model-level adapter to enhance downstream task performance. Additionally, \cite{yoshimura2023rawgment} proposed using noise-accounted raw augmentation to enhance recognition performance. Despite these advancements, most of these approaches often rely on simulating expert-tuning ISP stages, resulting in redundant parameters and increased computational complexity, which can adversely affect both accuracy and efficiency.

\paragraph{Image Enhancement based on RAW Data}
Many works have focused on enhancing images from RAW data. \cite{brooks2019unprocessing} introduced a method to transform sRGB images into synthetic RAW data and trained a neural network model for denoising. \cite{zou2023rawhdr} suggested separating challenging and easier regions and using dual intensity and global spatial guidance to reconstruct images from RAW images. \cite{9503334} proposed leveraging the green channel as a prior to jointly perform demosaicking and denoising on images, demonstrating the unique value of the green channel. However, these methods depend on paired datasets, which, even when artificially synthesized, often result in suboptimal data that may not be directly applicable to object detection tasks.

\section{Acknowledgement}
We thank Caizhi Zhu, Yinqiang Zheng, Xuanlong Yu, Zechao Hu, Hao Li, Zhengwei Yang, Song Ouyang, Jingyu Xu, Likai Tian, and Runqi Wang for their valuable support. This work was funded by the National Natural Science Foundation of China (Grant No. 62571379). The numerical calculations in this paper have been done on the supercomputing system in the Supercomputing Center of Wuhan University.

\bibliography{aaai2026}
\input{sup}

\end{document}

%% file: sup.tex
\urlstyle{rm} %
\def\UrlFont{\rm}  %
\frenchspacing  %
\setlength{\pdfpagewidth}{8.5in} %
\setlength{\pdfpageheight}{11in} %

\lstset{%
	basicstyle={\footnotesize\ttfamily},%
	numbers=left,numberstyle=\footnotesize,xleftmargin=2em,%
	aboveskip=0pt,belowskip=0pt,%
	showstringspaces=false,tabsize=2,breaklines=true}
\floatstyle{ruled}
\newfloat{listing}{tb}{lst}{}
\floatname{listing}{Listing}
\pdfinfo{
/TemplateVersion (2026.1)
}

\setcounter{secnumdepth}{0} %

\clearpage

\section{Appendix}
In this appendix, we present:
\begin{itemize}
    \item Segmentation Performance on the Real-World iPhone XSmax~\cite{li2024efficient} Dataset.
    \item Detection Performance on Synthetic Over-exposure and Dark Datasets. 
    \item Visualization, including additional comparison of our SimROD and DIAP~\cite{xu2023toward} on the detection dataset, comparison of SimROD and RAW-Adapter~\cite{raw_adapter} on the detection dataset, and comparison of SimROD and RAW-Adapter~\cite{raw_adapter} on the segmentation dataset.
    \item More analysis of our Global Gamma Enhancement(GGE), including a brief introduction to the Gamma Transformation, hyperparameters sensitivity analysis, ablation study of GGE and comparision between sRGB and RAW. 
    \item We compared the inference time with DIAP~\cite{xu2023toward} and RAW-Adapter~\cite{raw_adapter}, further demonstrating the efficiency of our approach.
\end{itemize}

\begin{figure*}[!t]
    \label{fig:gamma_sens}
    \centering
    \begin{minipage}{\linewidth}
        \begin{subfigure}{0.45\textwidth}
            \centering
            \includegraphics[width=1.0\linewidth]{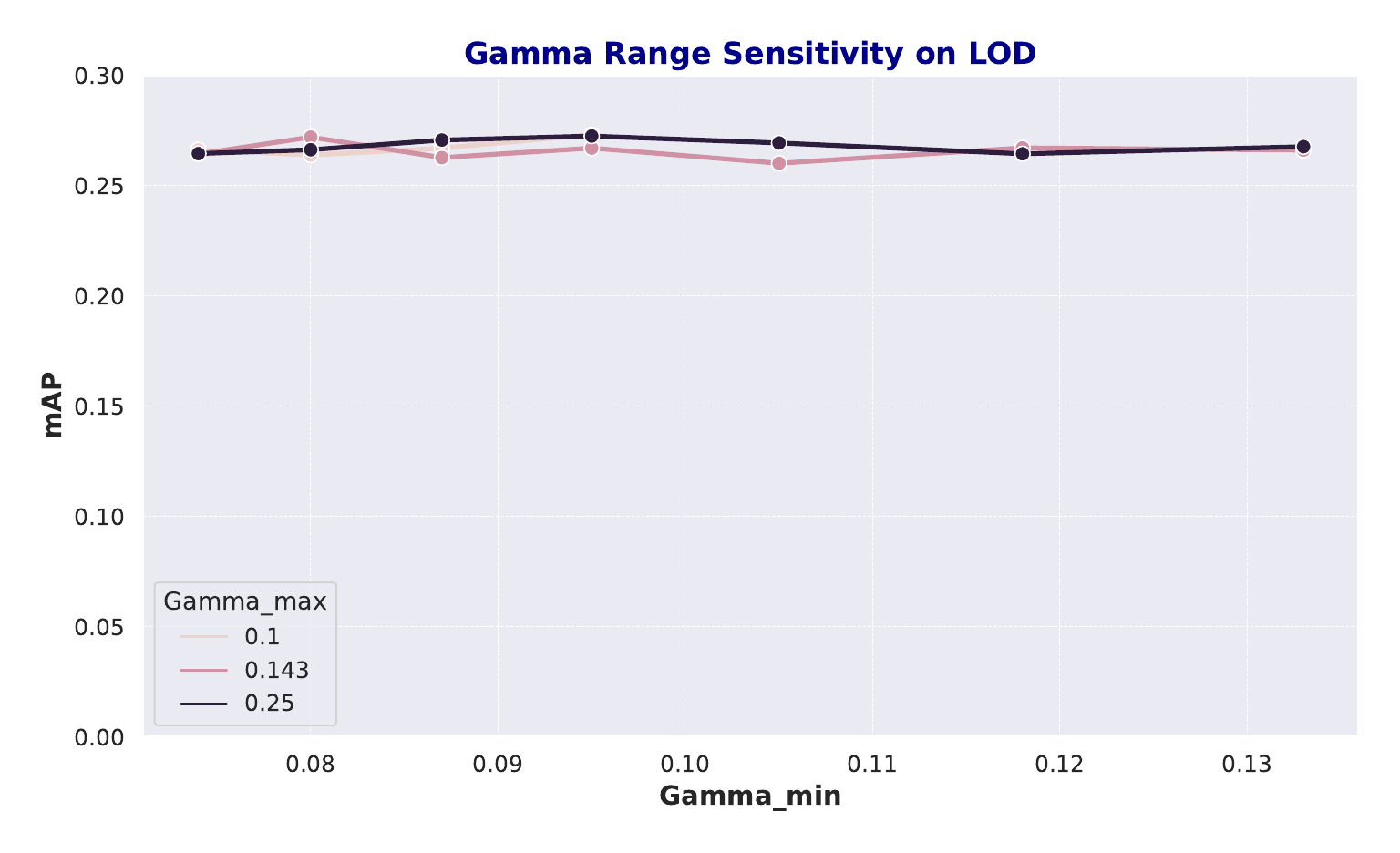}
            \caption{\textbf{Effect of Gamma Range on LOD~\cite{hong2021crafting}.}}
            \label{fig:gamma_lod}
        \end{subfigure}
        \begin{subfigure}{0.45\textwidth}
            \centering
            \includegraphics[width=1.0\linewidth]{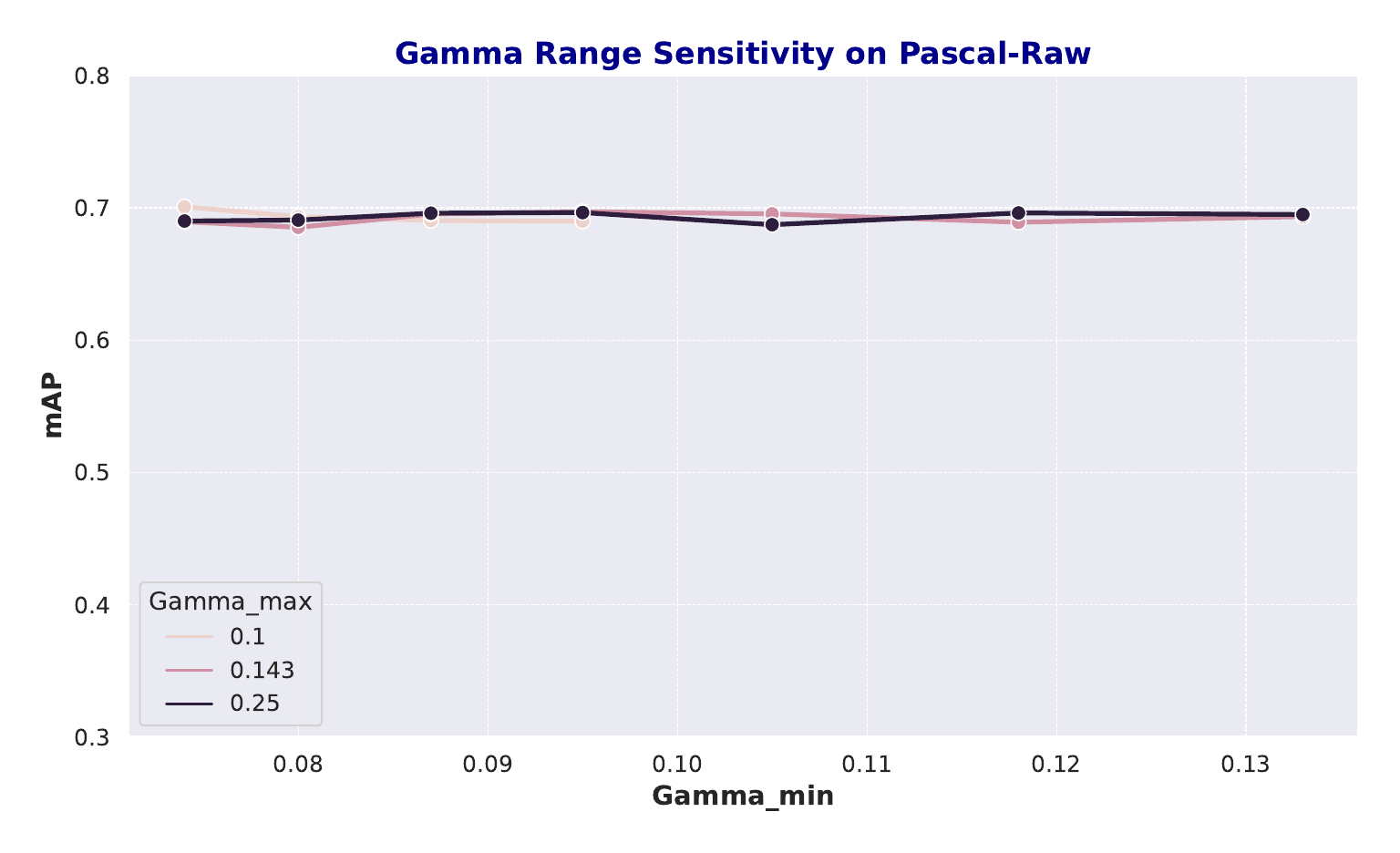}
            \caption{\textbf{Effect of Gamma Range on Pascal-Raw~\cite{omid2014pascalraw}.}}
            \label{fig:gamma_pas}
        \end{subfigure}
    \end{minipage}
    \caption{The plots illustrate the sensitivity of mAP to varying \textit{gamma\_min}($\gamma_{min}$) and \textit{gamma\_max}($\gamma_{max}$) defined in our GGE. The results show minimal performance variation across different gamma ranges, indicating robust detection performance within the tested parameter bounds.}
\end{figure*}

\subsection{Segmentation on Real-World iPhone XSmax Dataset}
\label{sec_iphone}
The iPhone XSmax~\cite{li2024efficient} is a real-world RAW semantic segmentation dataset that consists of 1153 RAW images with their corresponding semantic labels, where 806 images are set as training set and the other 347 images are set as evaluation set. Following Raw-adapter~\cite{raw_adapter}, we adopt Segformer\cite{xie2021segformer} framework with MIT-B5 backbone, training iterations are set to 20000, and other settings are same as ADE 20K RAW~\cite{raw_adapter}'s setting, as defined by ~\cite{raw_adapter}. As shown in Table~\ref{tab:iphone}, the SimROD could also achieve superior results with its parameter efficiency. 

\begin{table}[htbp]
  \centering
  \caption{\textbf{Segmentation Performance on the Real-World iPhone XSmax~\cite{li2024efficient} Dataset.} Bold number indicates the best result.}
    \begin{tabular}{ccc}
    \toprule
    Method & mIOU & Add. Params.(M) \\
    \midrule
    Demosacing & 59.81  & 0.00  \\
    DefaultISP & 61.82  & 0.00  \\
    Karaimer et Brown & 60.41  & 0.00  \\
    LiteISP & 59.84  & 0.00  \\
    Dirty-Pixel & 61.85  & 4.28  \\
    RAW-Adapter & 62.45  & 0.30  \\
    Our SimROD & \textbf{63.13}  & 0.003  \\
    \bottomrule
    \end{tabular}%
  \label{tab:iphone}%
\end{table}%

\subsection{Detection Performance on Synthetic Over-exposure and Dark Datasets}
Following ~\cite{raw_adapter}, we also report results on the Over-exp and Dark versions in Tab.~\ref{tab:re_oe_low}, where SimROD continues to outperform existing methods by a large margin.

\begin{table}[htbp]
  \centering
  \caption{Over-exposure and dark version Pascal-Raw.}
    \begin{tabular}{ccccc}
    \toprule
    Method & Over-exposure & Dark \\
    \midrule
    Dirty-Pixel [8,9] & 88.0  & 80.8 \\
    RAW-Adapter & 88.7    & 82.5  \\
    SimROD & \textbf{89.4} & \textbf{84.1} \\
    \bottomrule
    \end{tabular}%
  \label{tab:re_oe_low}%
\end{table}%

\subsection{Visualization}
\begin{figure*}[ht]
    \centering
    \begin{minipage}{\textwidth}
                
        \begin{subfigure}{\textwidth}
            \centering
            \textbf{\footnotesize RAW Images w. GT}
        \end{subfigure}

        \begin{subfigure}{0.245\textwidth}
            \includegraphics[width=\linewidth]{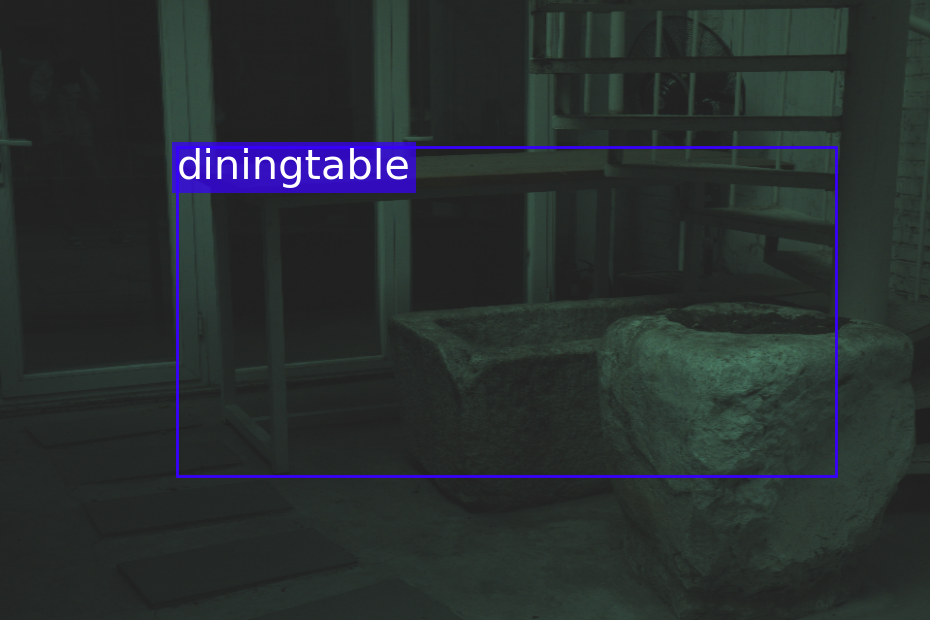}
        \end{subfigure}
         \begin{subfigure}{0.245\textwidth}
            \includegraphics[width=\linewidth]{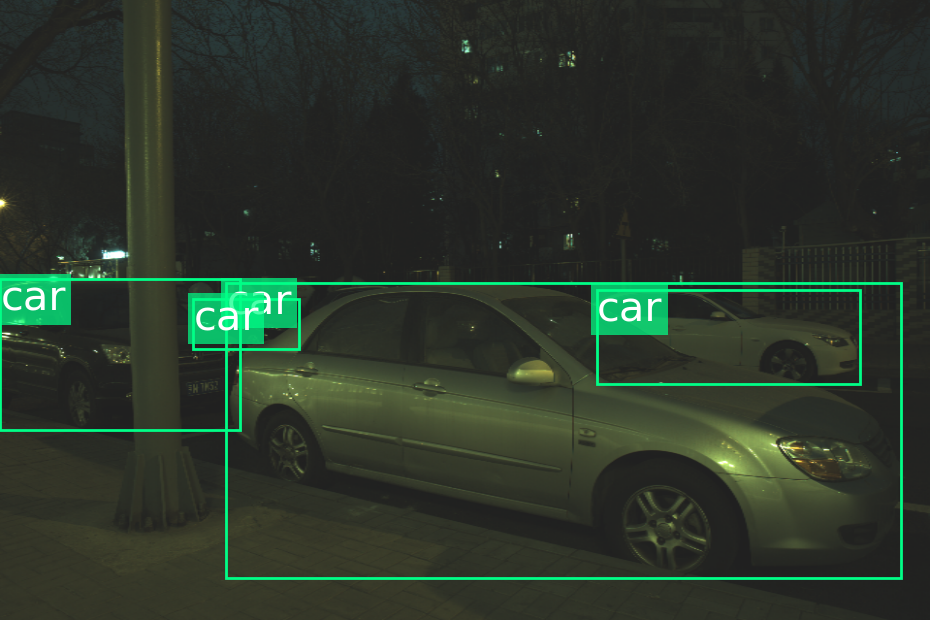}
        \end{subfigure}
         \begin{subfigure}{0.245\textwidth}
            \includegraphics[width=\linewidth]{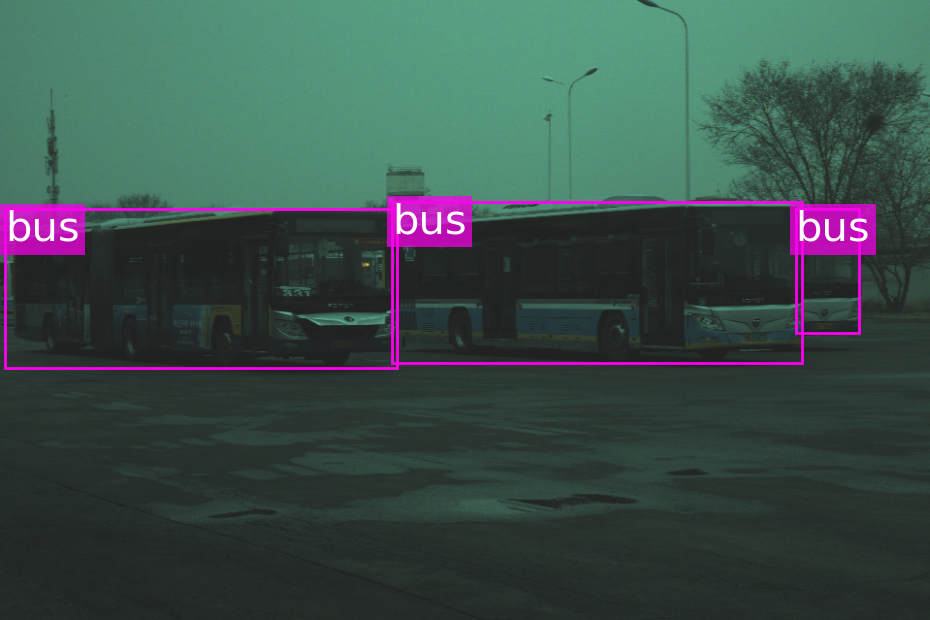}
        \end{subfigure}
        \begin{subfigure}{0.245\textwidth}
            \includegraphics[width=\linewidth]{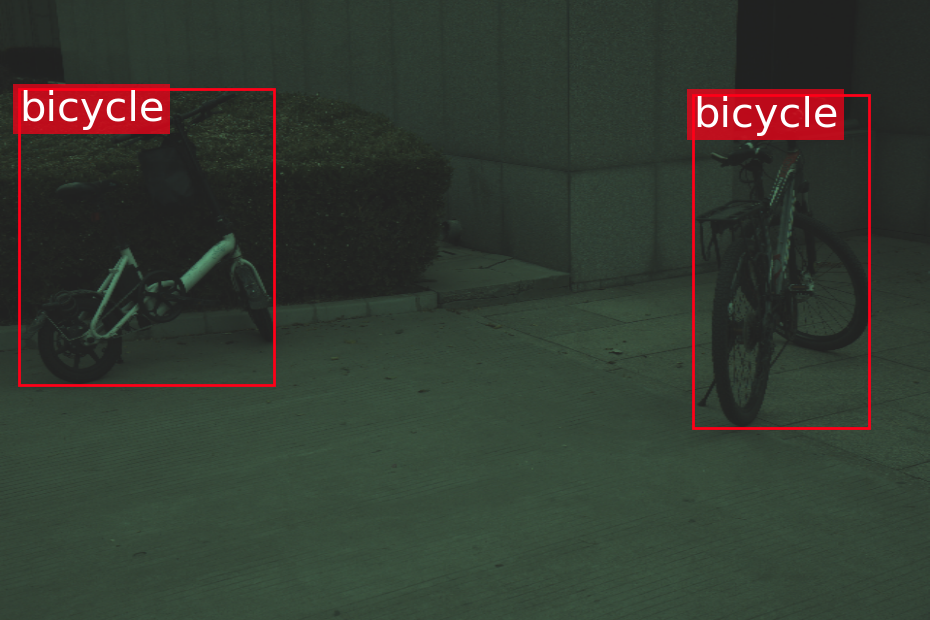}
        \end{subfigure}
        
        \begin{subfigure}{\textwidth}
            \centering
            \textbf{\footnotesize RAW Distribution}
        \end{subfigure}

        \begin{subfigure}{0.245\textwidth}
            \includegraphics[width=\linewidth]{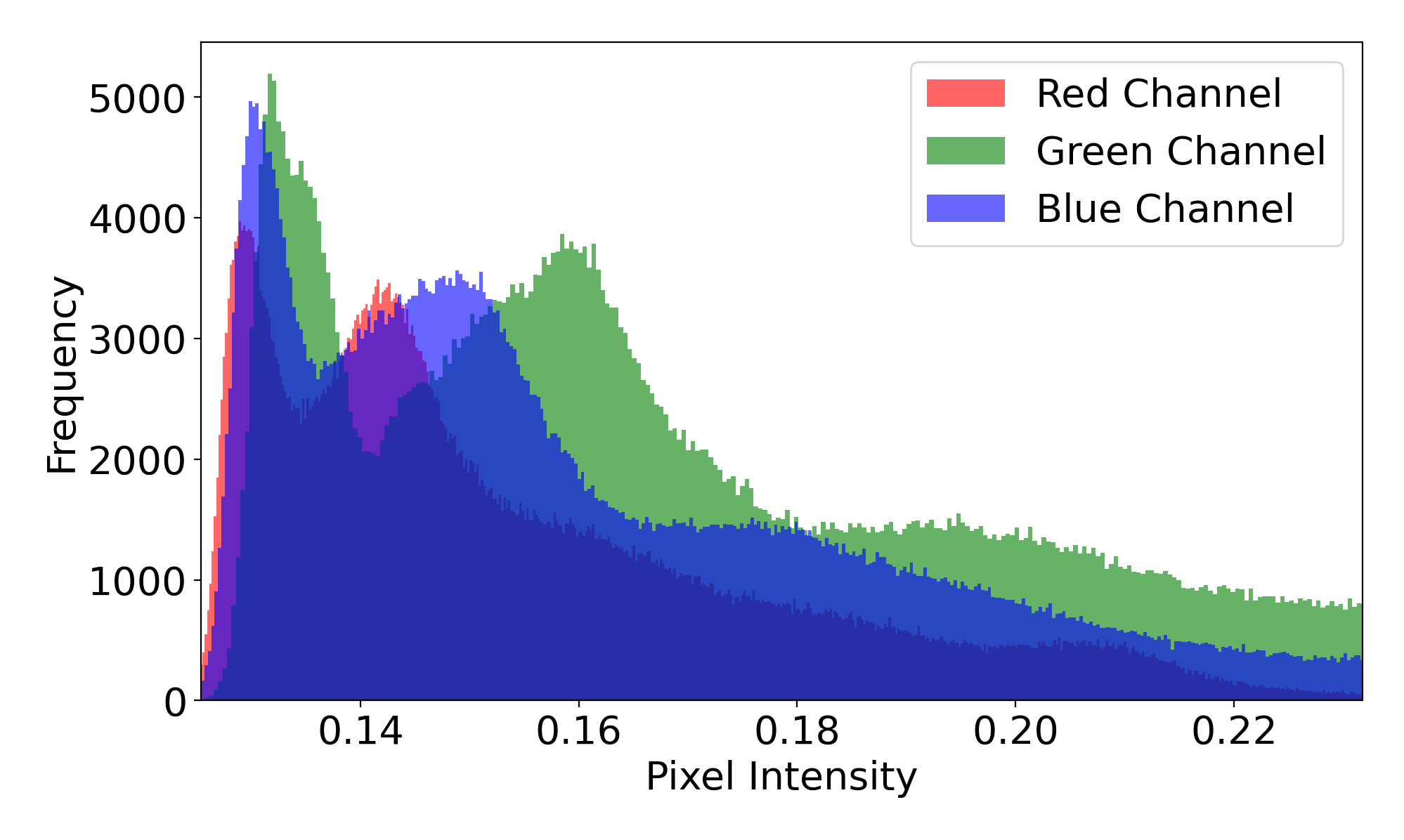}
        \end{subfigure}
         \begin{subfigure}{0.245\textwidth}
            \includegraphics[width=\linewidth]{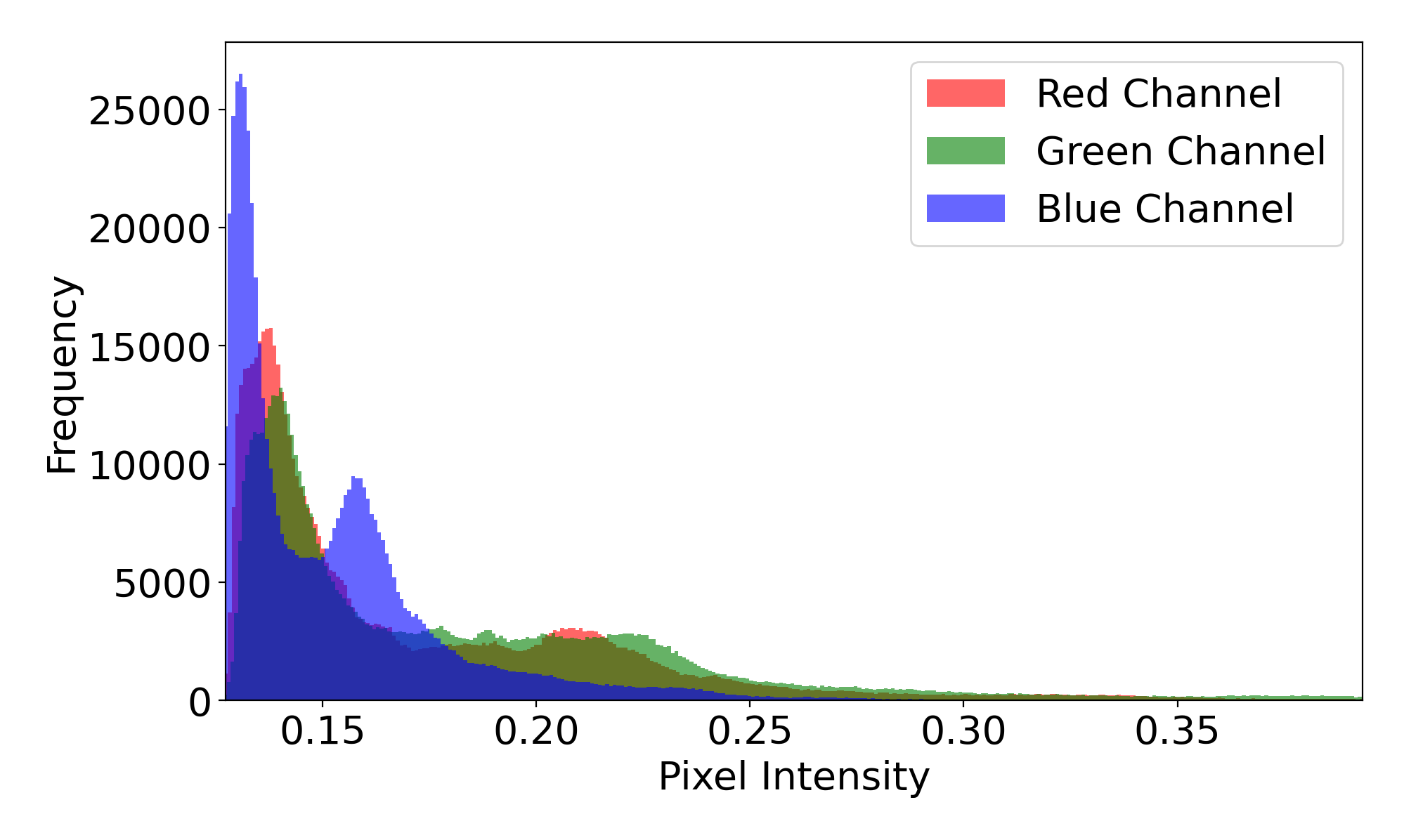}
        \end{subfigure}
         \begin{subfigure}{0.245\textwidth}
            \includegraphics[width=\linewidth]{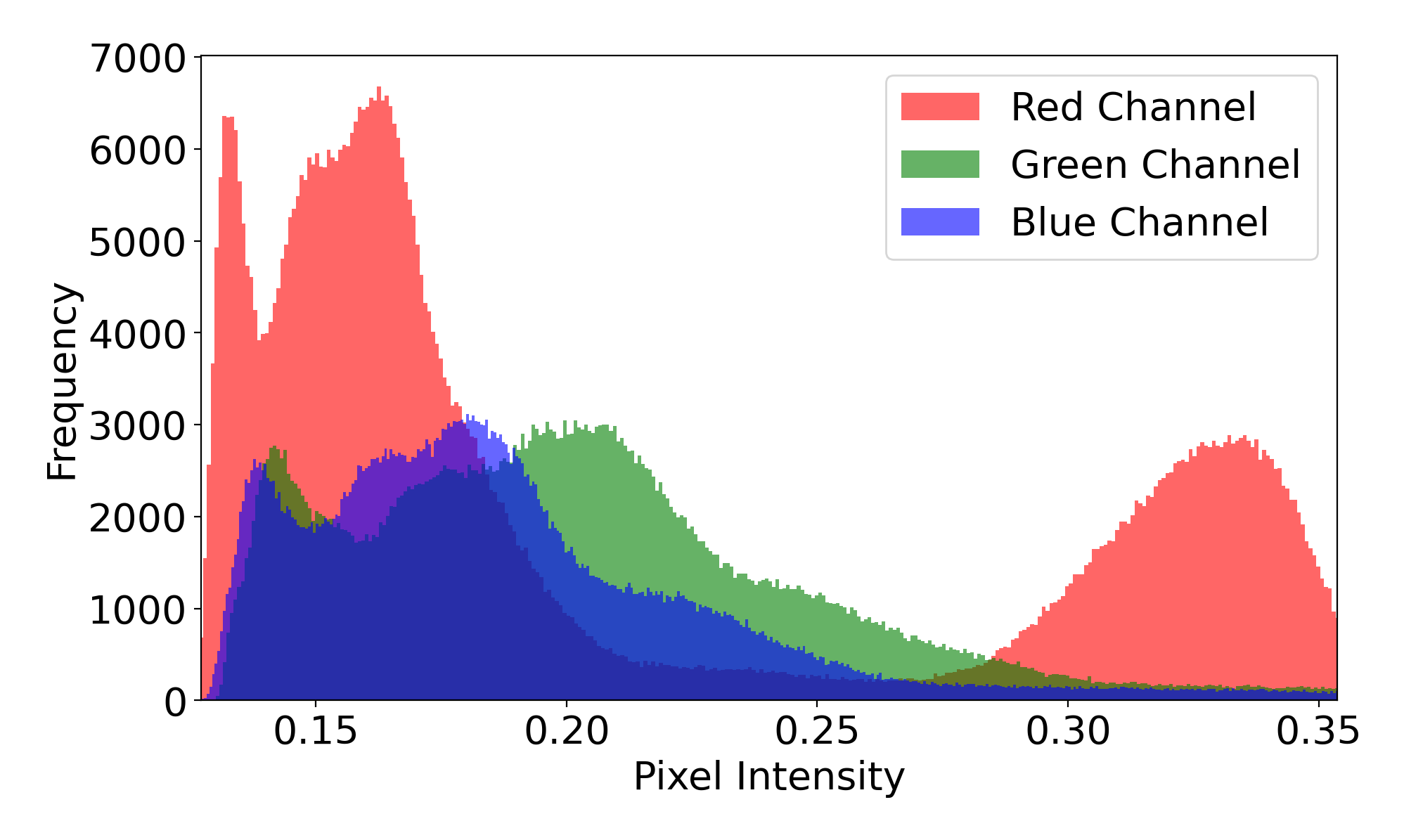}
        \end{subfigure}
        \begin{subfigure}{0.245\textwidth}
            \includegraphics[width=\linewidth]{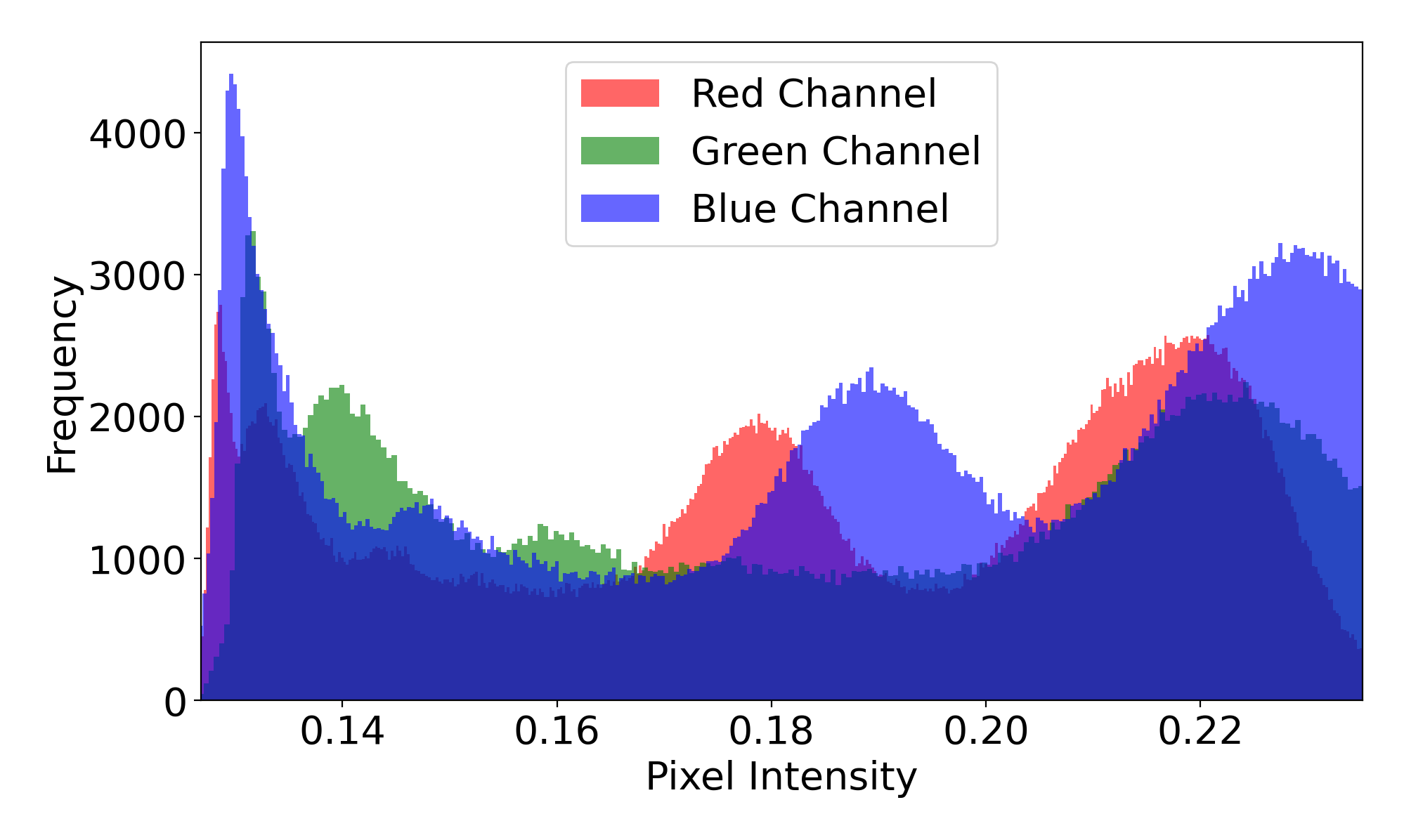}
        \end{subfigure}

        \begin{subfigure}{\textwidth}
            \centering
            \textbf{\footnotesize DIAP~\cite{xu2023toward} Det.}
        \end{subfigure}

        \begin{subfigure}{0.245\textwidth}
            \includegraphics[width=\linewidth]{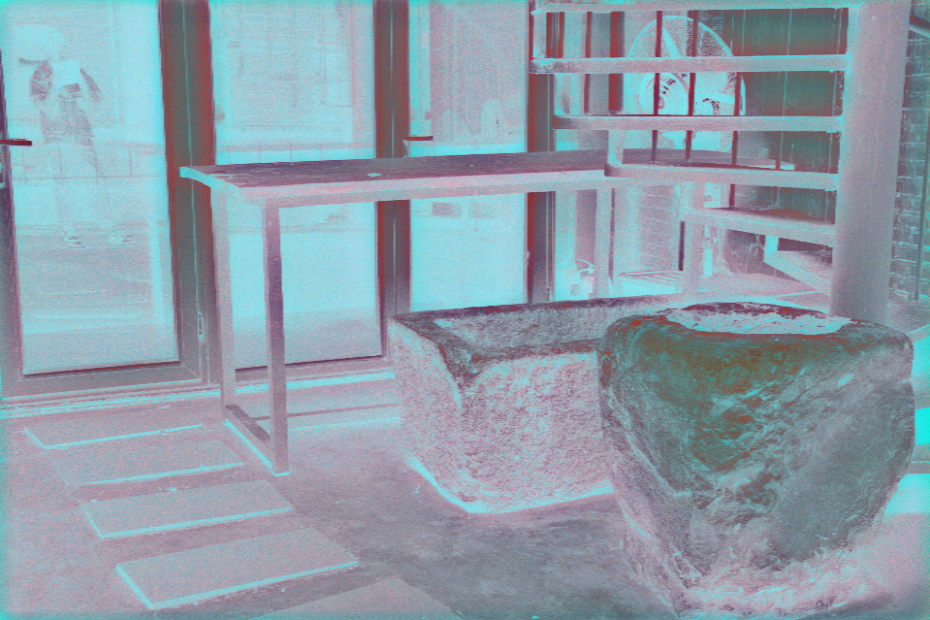}
        \end{subfigure}
         \begin{subfigure}{0.245\textwidth}
            \includegraphics[width=\linewidth]{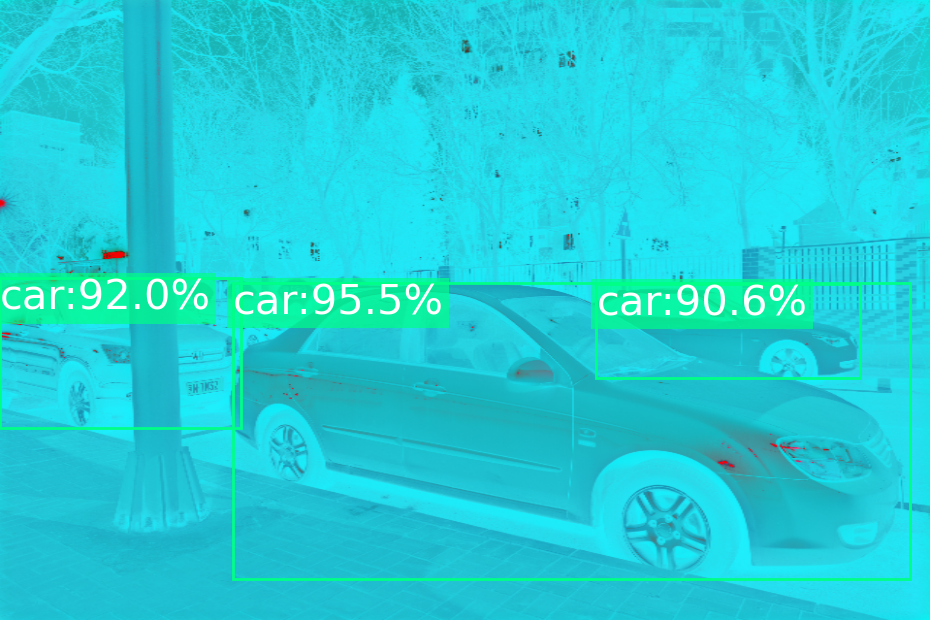}
        \end{subfigure}
         \begin{subfigure}{0.245\textwidth}
            \includegraphics[width=\linewidth]{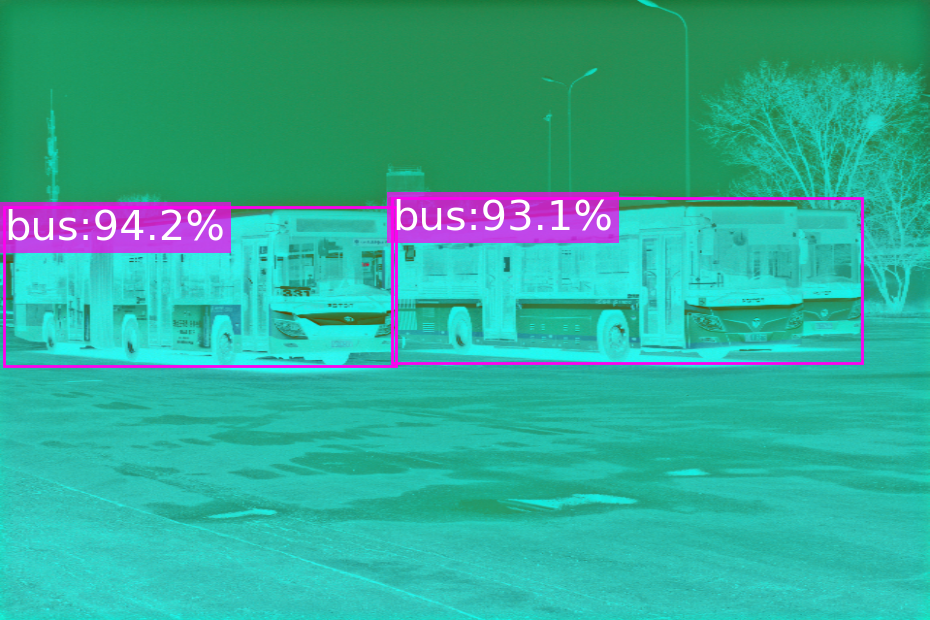}
        \end{subfigure}
        \begin{subfigure}{0.245\textwidth}
            \includegraphics[width=\linewidth]{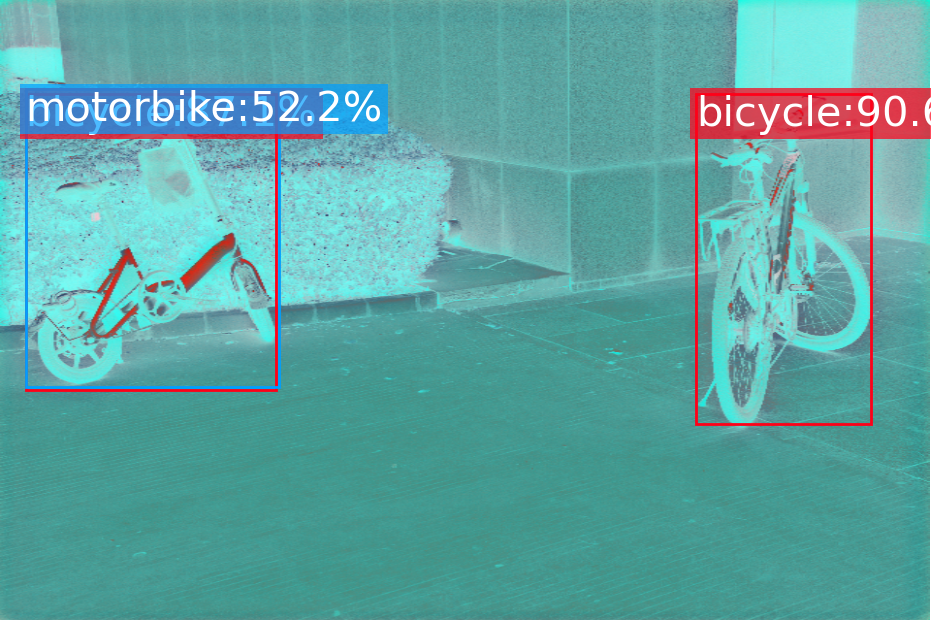}
        \end{subfigure}

        \begin{subfigure}{\textwidth}
            \centering
            \textbf{\footnotesize Distribution after DIAP~\cite{xu2023toward}}
        \end{subfigure}

        \begin{subfigure}{0.245\textwidth}
            \includegraphics[width=\linewidth]{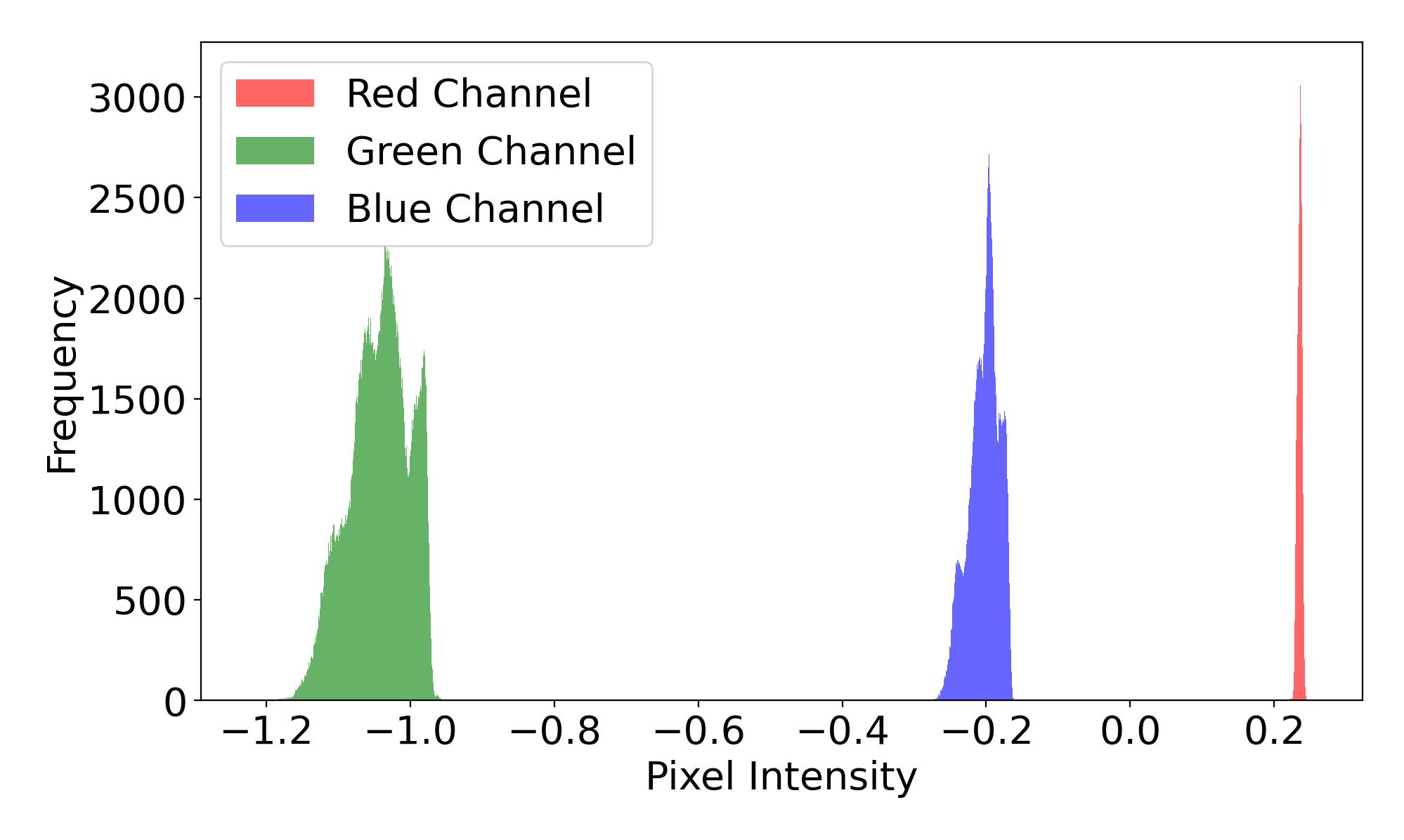}
        \end{subfigure}
         \begin{subfigure}{0.245\textwidth}
            \includegraphics[width=\linewidth]{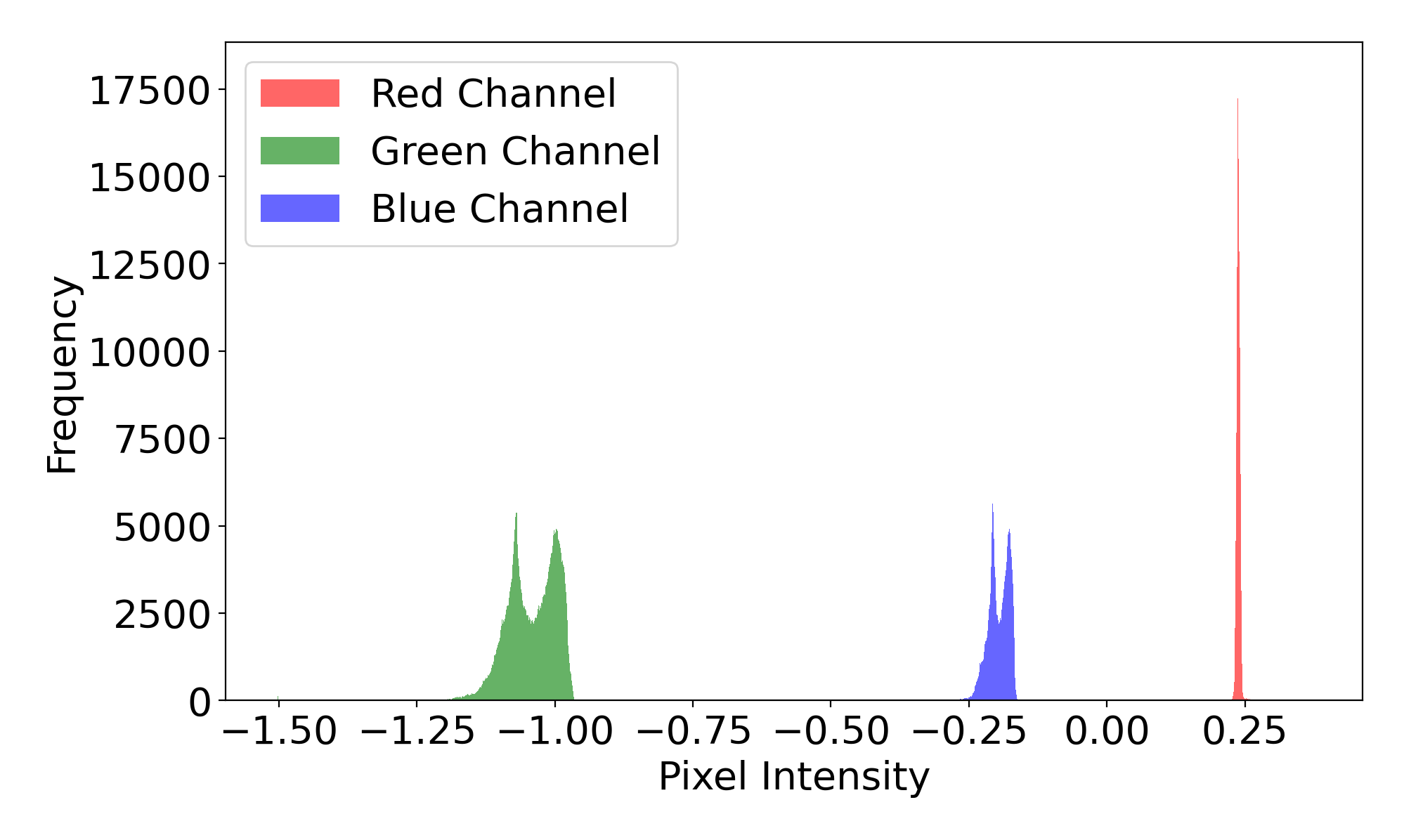}
        \end{subfigure}
         \begin{subfigure}{0.245\textwidth}
            \includegraphics[width=\linewidth]{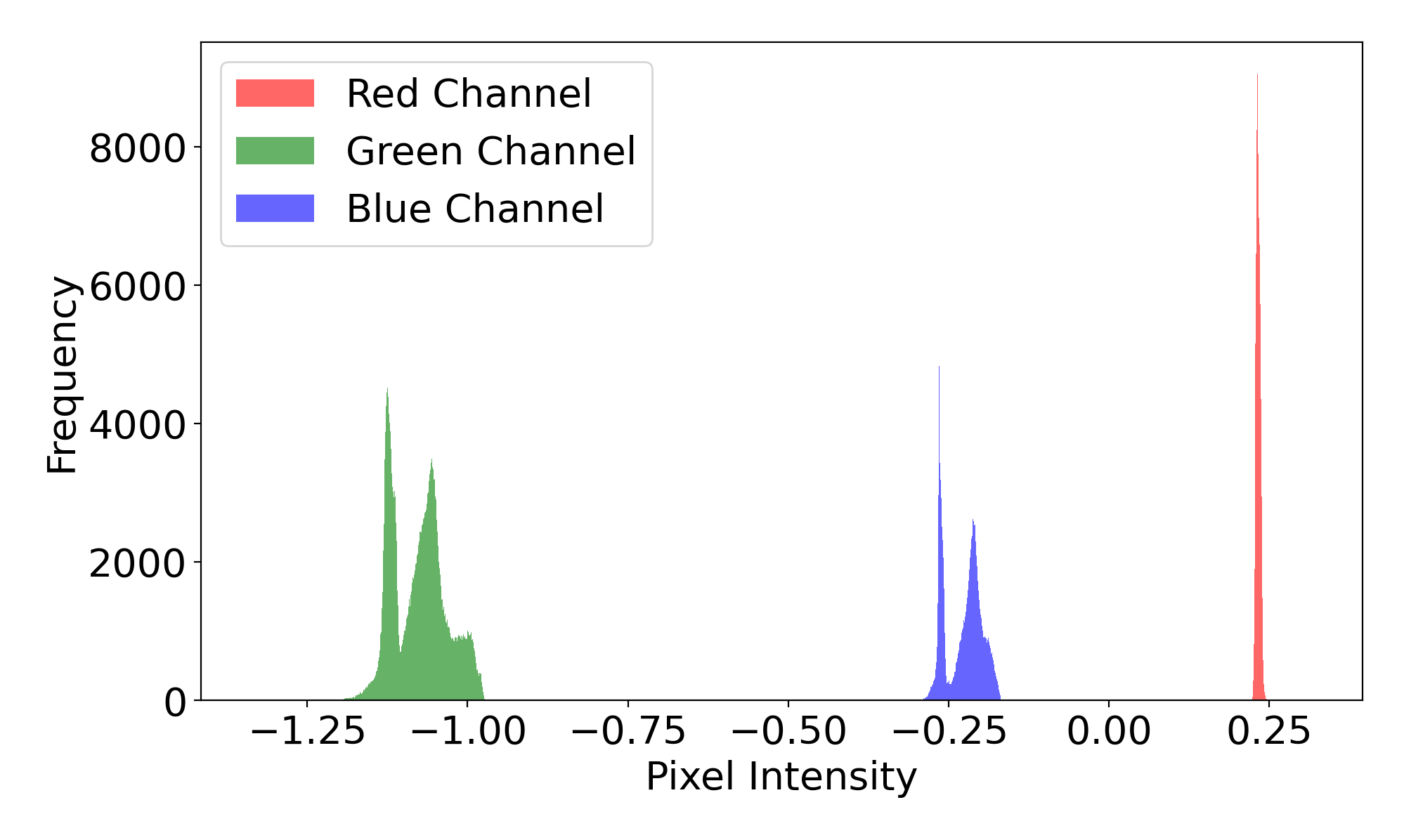}
        \end{subfigure}
        \begin{subfigure}{0.245\textwidth}
            \includegraphics[width=\linewidth]{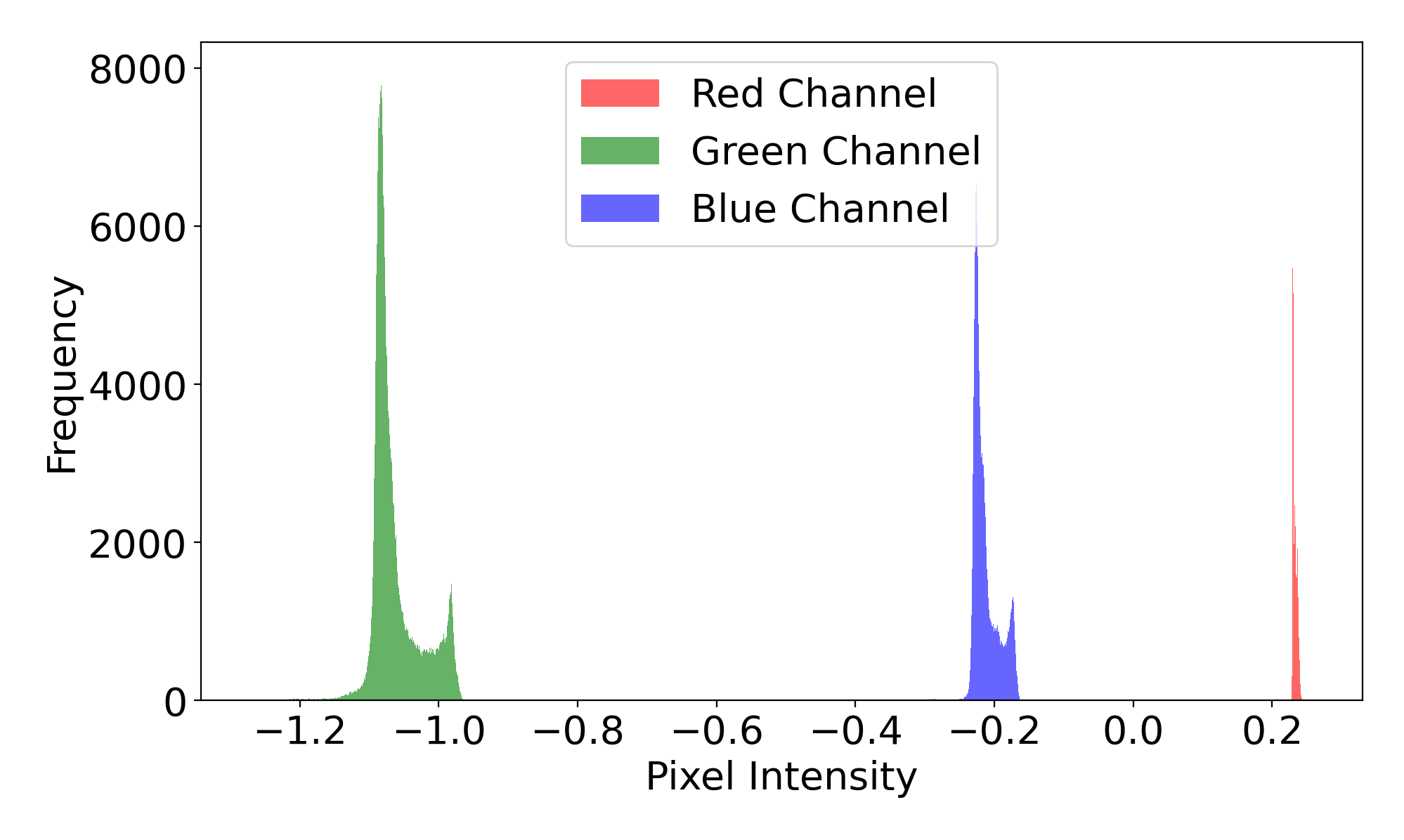}
        \end{subfigure}

        \begin{subfigure}{\textwidth}
            \centering
            \textbf{\footnotesize Our SimROD Det.}
        \end{subfigure}

        \begin{subfigure}{0.245\textwidth}
            \includegraphics[width=\linewidth]{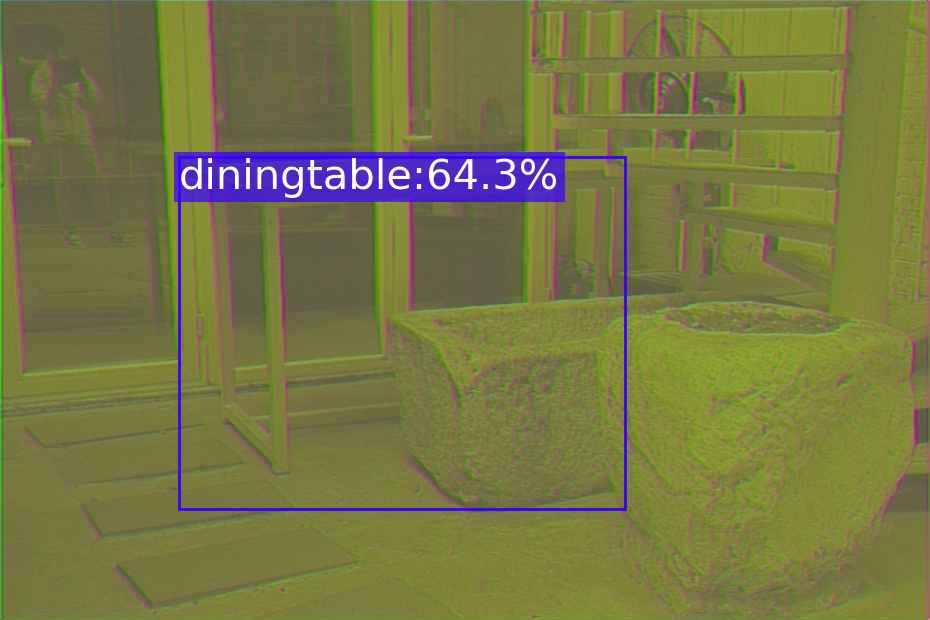}
        \end{subfigure}
         \begin{subfigure}{0.245\textwidth}
            \includegraphics[width=\linewidth]{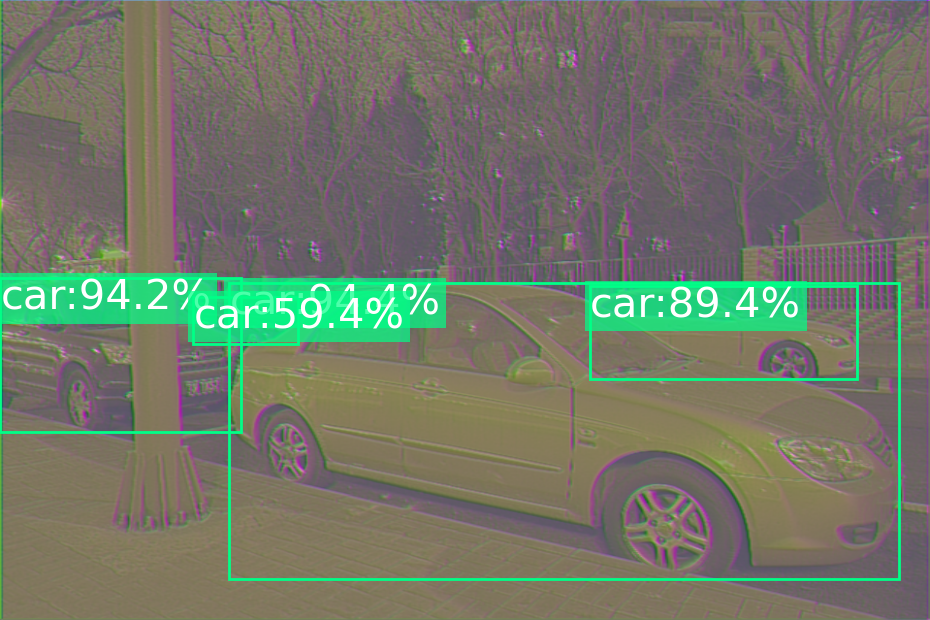}
        \end{subfigure}
         \begin{subfigure}{0.245\textwidth}
            \includegraphics[width=\linewidth]{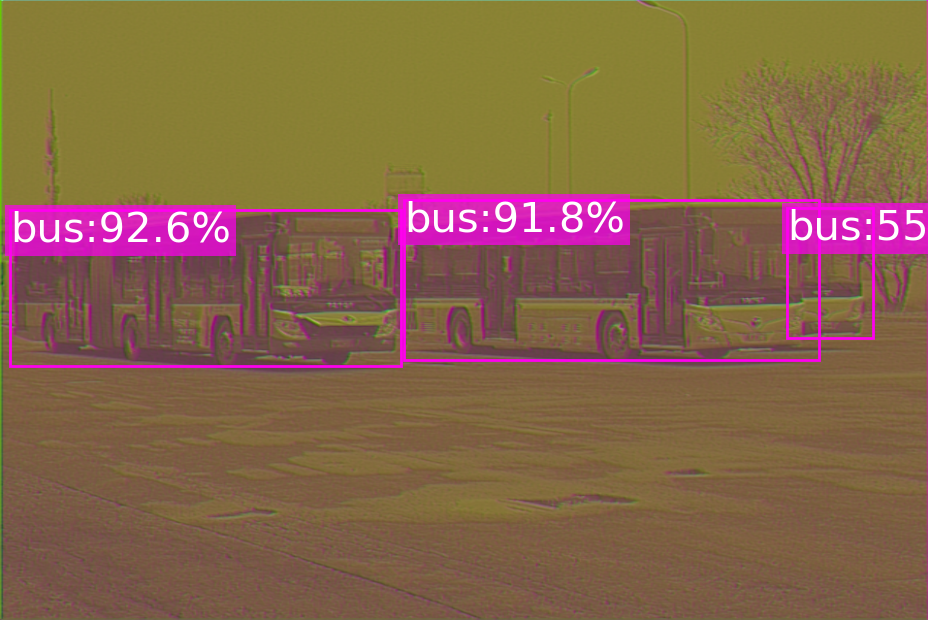}
        \end{subfigure}
        \begin{subfigure}{0.245\textwidth}
            \includegraphics[width=\linewidth]{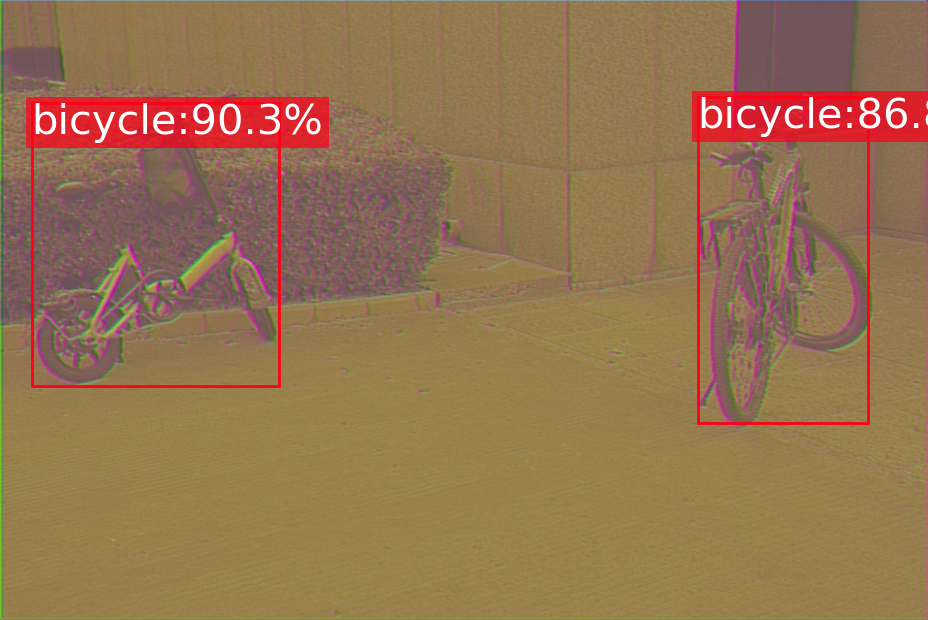}
        \end{subfigure}

        \begin{subfigure}{\textwidth}
            \centering
            \textbf{\footnotesize Distribution after Our SimROD}
        \end{subfigure}

        \begin{subfigure}{0.245\textwidth}
            \includegraphics[width=\linewidth]{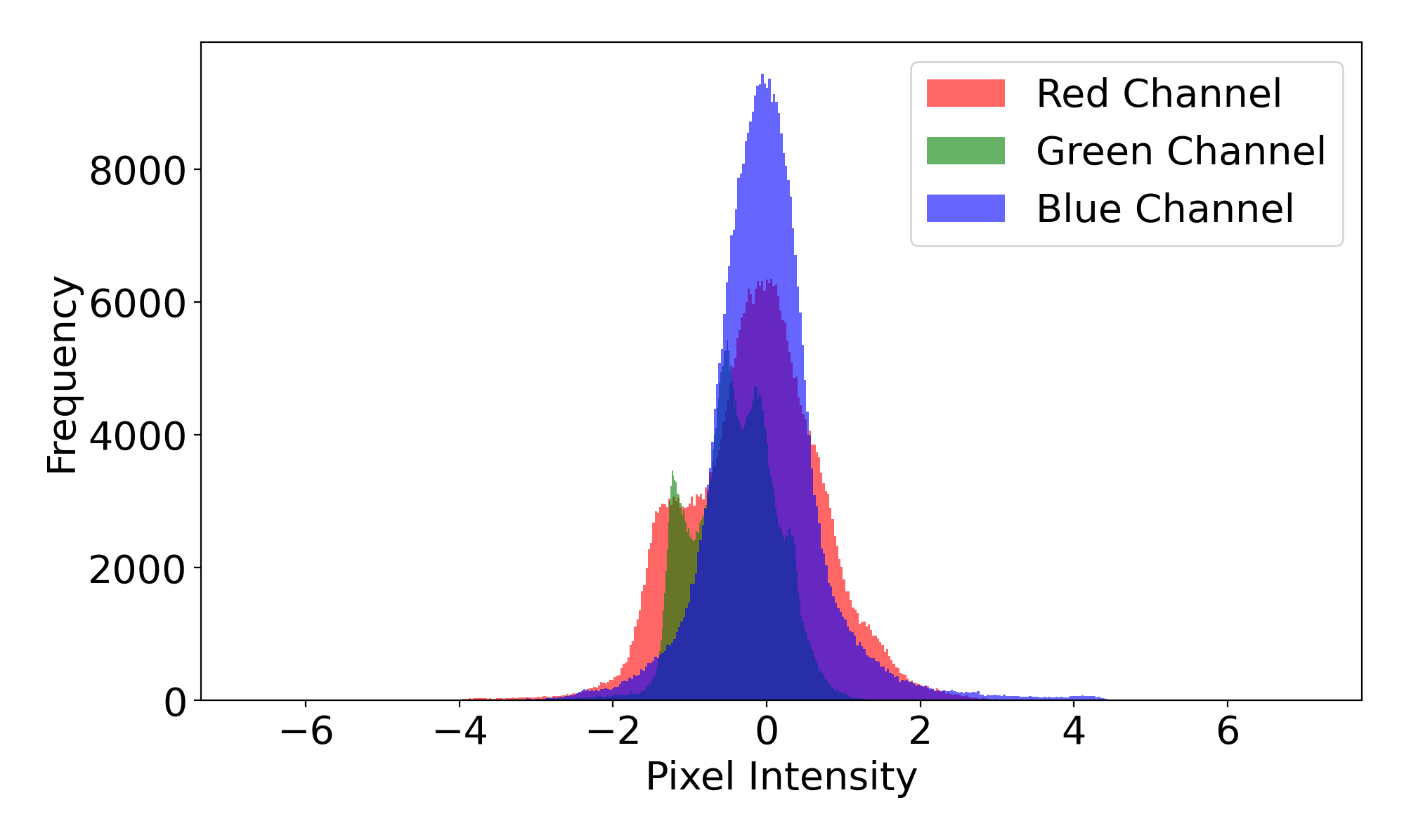}
        \end{subfigure}
         \begin{subfigure}{0.245\textwidth}
            \includegraphics[width=\linewidth]{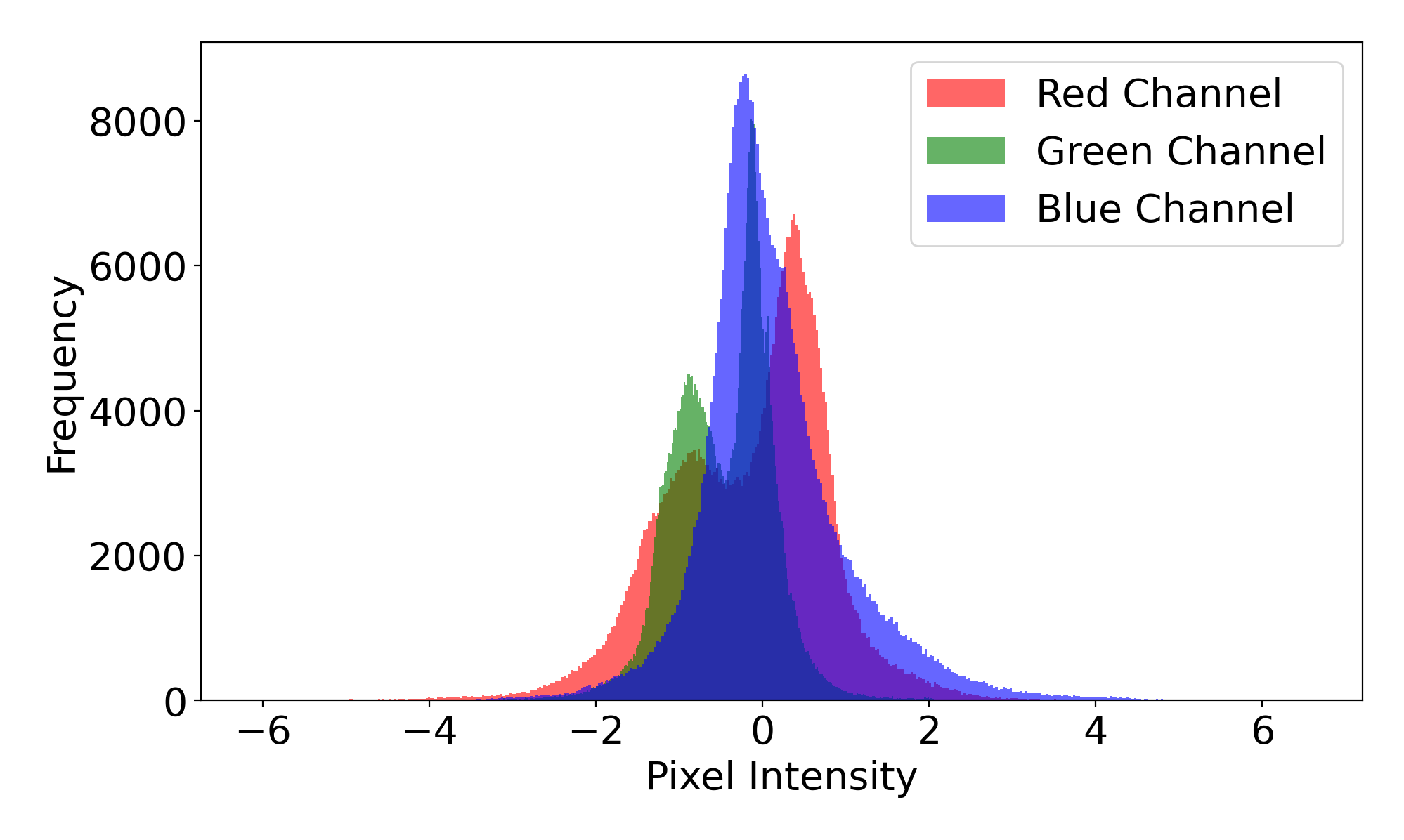}
        \end{subfigure}
         \begin{subfigure}{0.245\textwidth}
            \includegraphics[width=\linewidth]{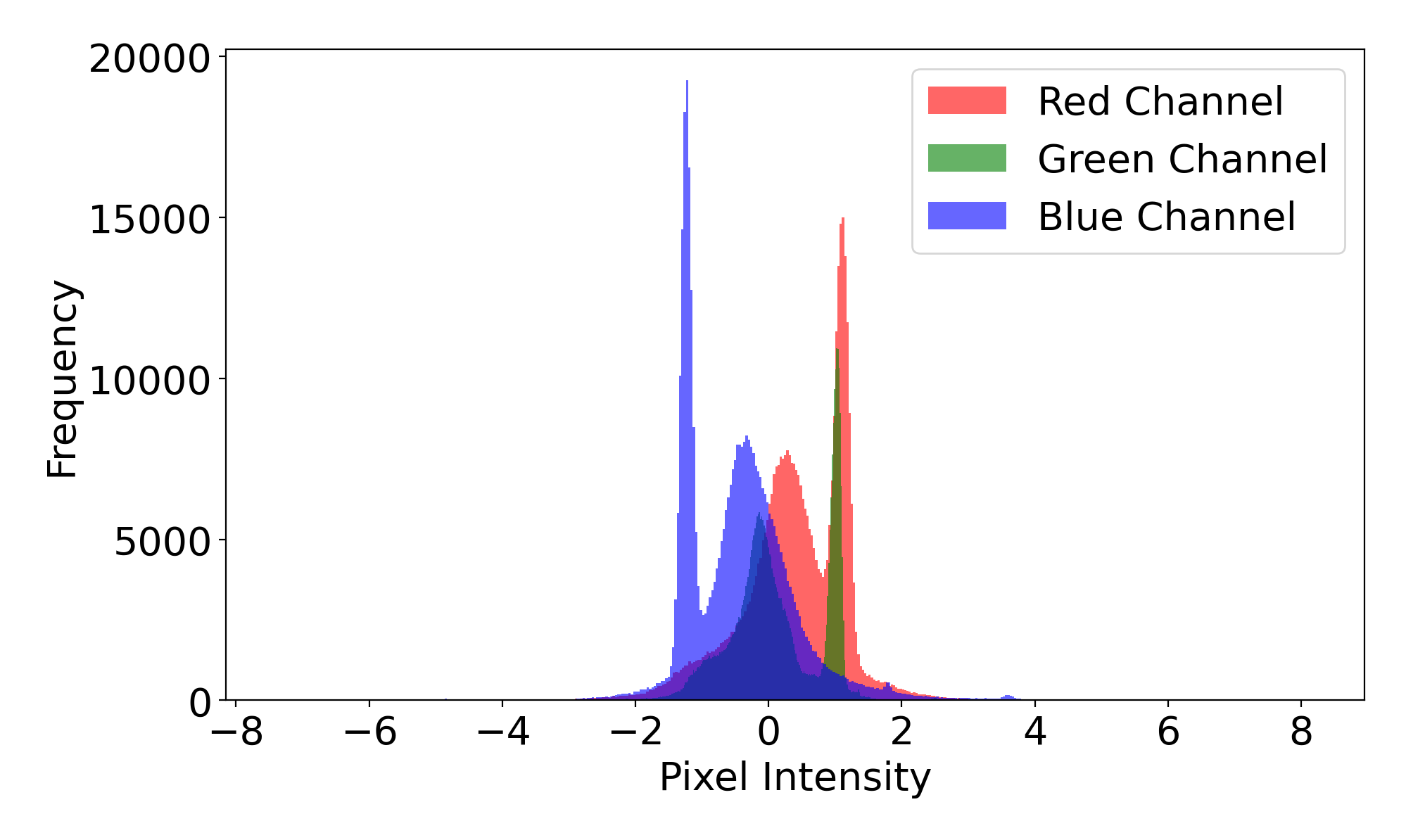}
        \end{subfigure}
        \begin{subfigure}{0.245\textwidth}
            \includegraphics[width=\linewidth]{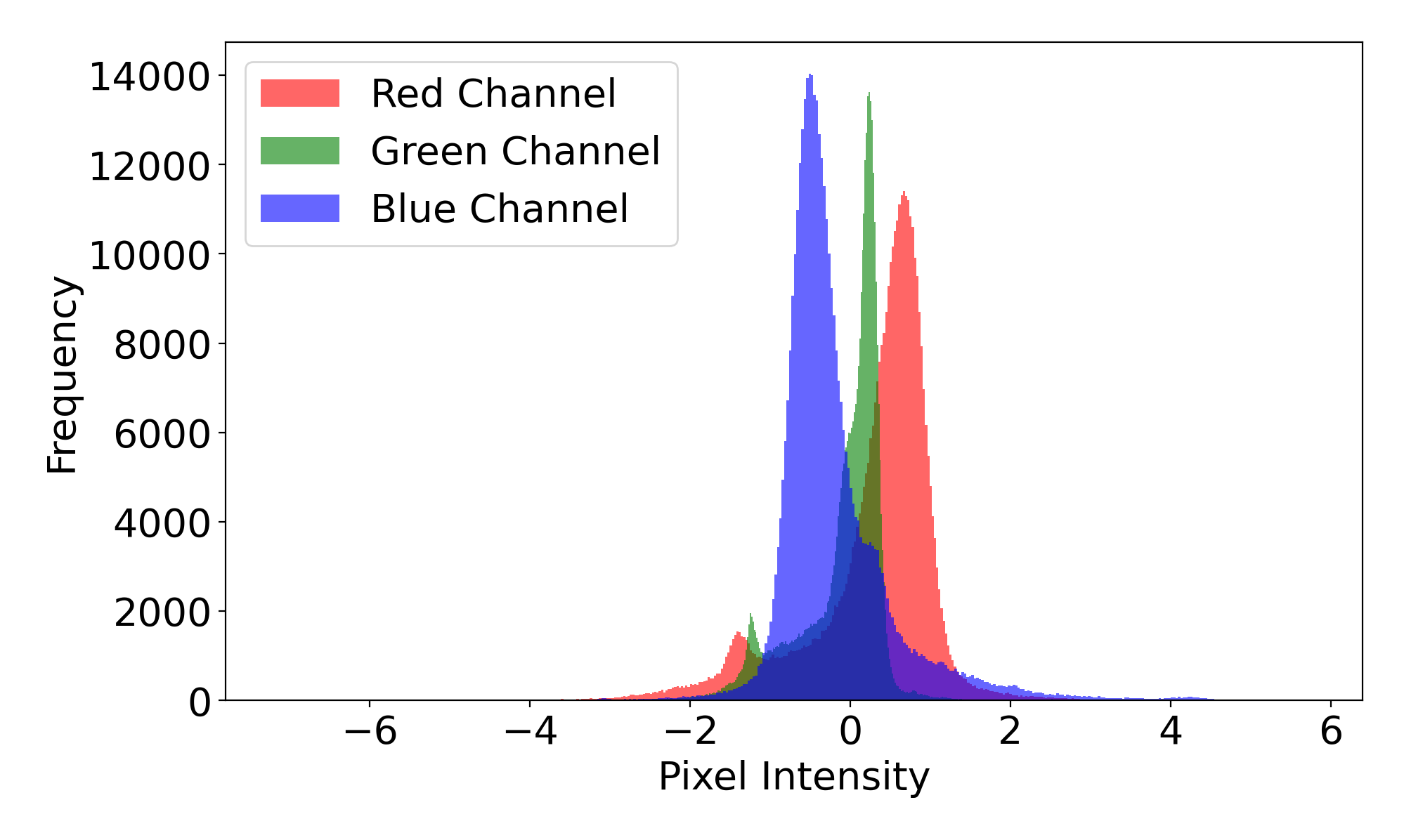}
        \end{subfigure}

    \end{minipage}
    \caption{\textbf{Visualization on the Object Detection Dataset.} We present detection and distribution results processed with DIAP~\cite{xu2023toward} and our SimROD. The figure includes RAW images with detection annotations, detection results, and RGB pixel distributions for both methods. The pixel distribution of the enhanced images from our SimROD more closely approximates a normal distribution, which facilitates more efficient feature learning for neural networks.}
    \label{fig:vis_sup_diap}
\end{figure*}

Figure \ref{fig:vis_sup_diap} provides a comprehensive visualization of detection results and channel distributions for both the DIAP~\cite{xu2023toward} and our SimROD across various scenes. A clear pattern emerges from these visualizations: our method consistently produces more accurate and stable detection results compared to DIAP~\cite{xu2023toward}. Notably, the pixel distribution of the enhanced images more closely approximates a normal distribution. This characteristic facilitates more efficient feature learning for neural networks, reducing the impact of outliers and thus improving detection accuracy. 

\begin{figure*}[ht]
    \centering
    \begin{minipage}{\textwidth}

        \begin{subfigure}{0.32\textwidth}
            \includegraphics[width=\linewidth]{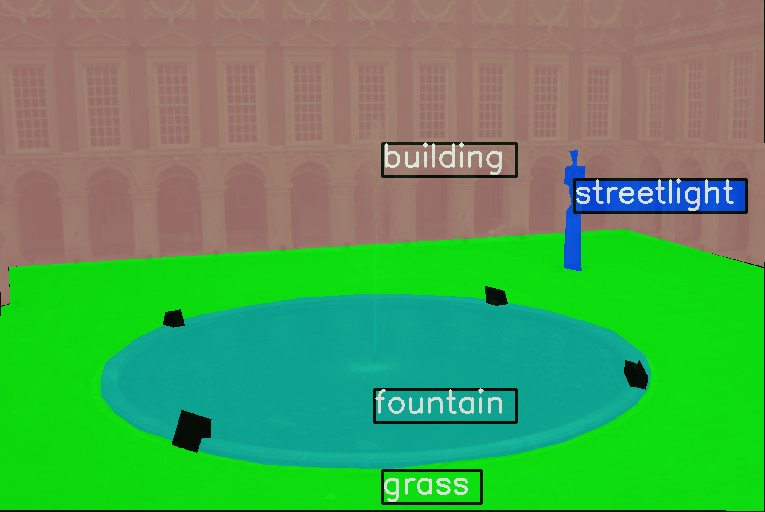}
        \end{subfigure}
         \begin{subfigure}{0.32\textwidth}
            \includegraphics[width=\linewidth]{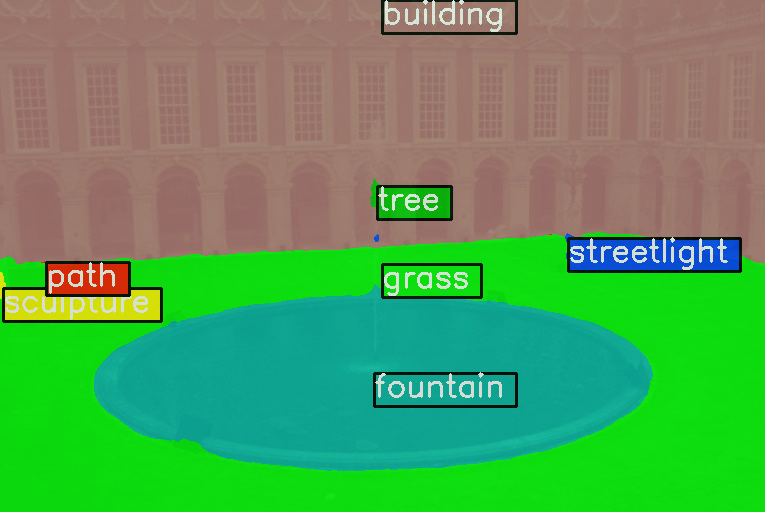}
        \end{subfigure}
         \begin{subfigure}{0.32\textwidth}
            \includegraphics[width=\linewidth]{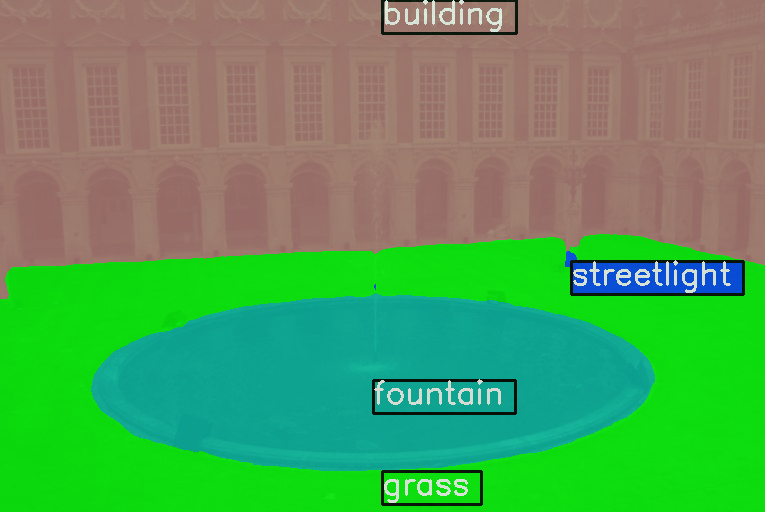}
        \end{subfigure}

        \begin{subfigure}{0.32\textwidth}
            \includegraphics[width=\linewidth]{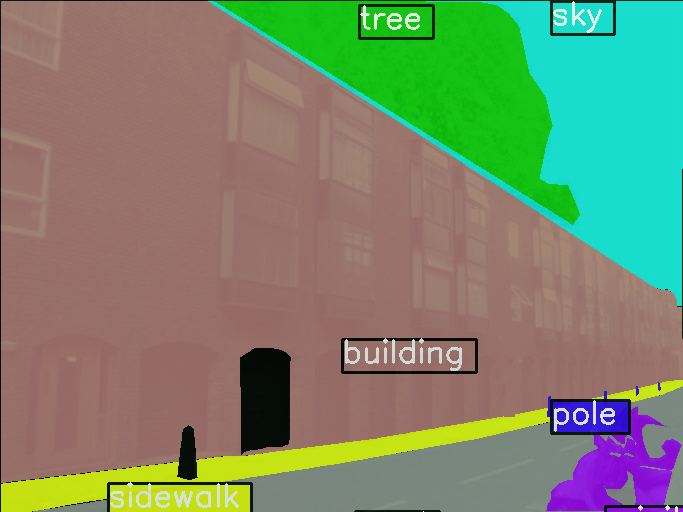}
        \end{subfigure}
         \begin{subfigure}{0.32\textwidth}
            \includegraphics[width=\linewidth]{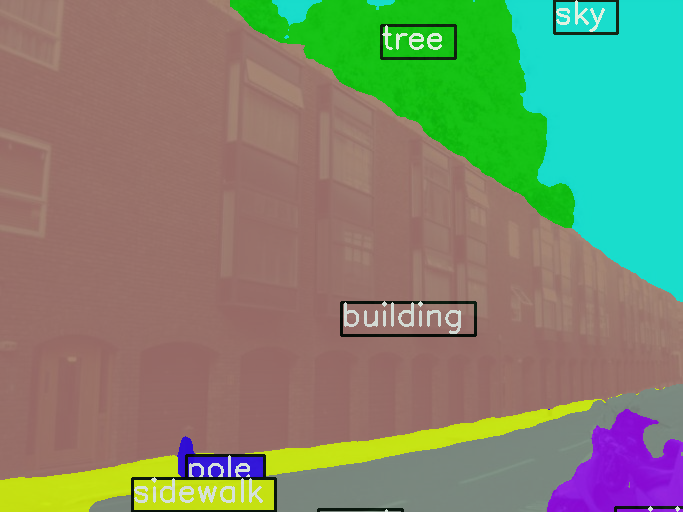}
        \end{subfigure}
         \begin{subfigure}{0.32\textwidth}
            \includegraphics[width=\linewidth]{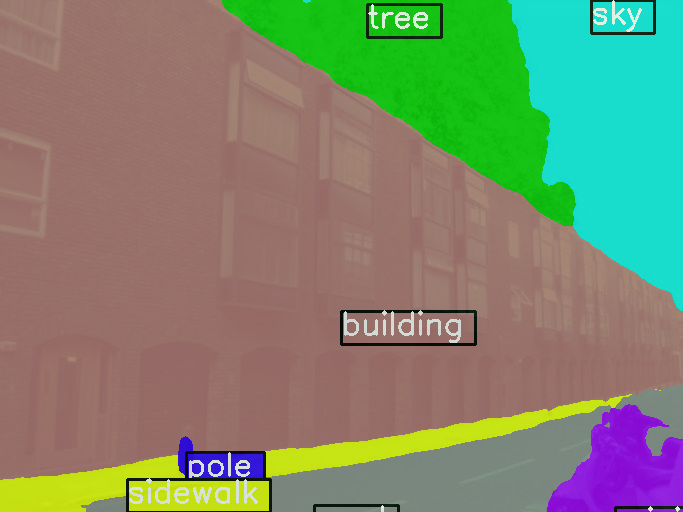}
        \end{subfigure}

        \begin{subfigure}{0.32\textwidth}
            \includegraphics[width=\linewidth]{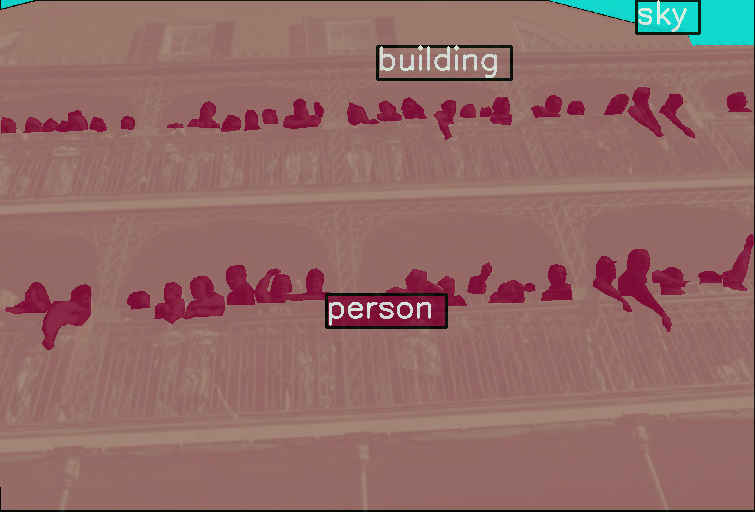}
        \end{subfigure}
         \begin{subfigure}{0.32\textwidth}
            \includegraphics[width=\linewidth]{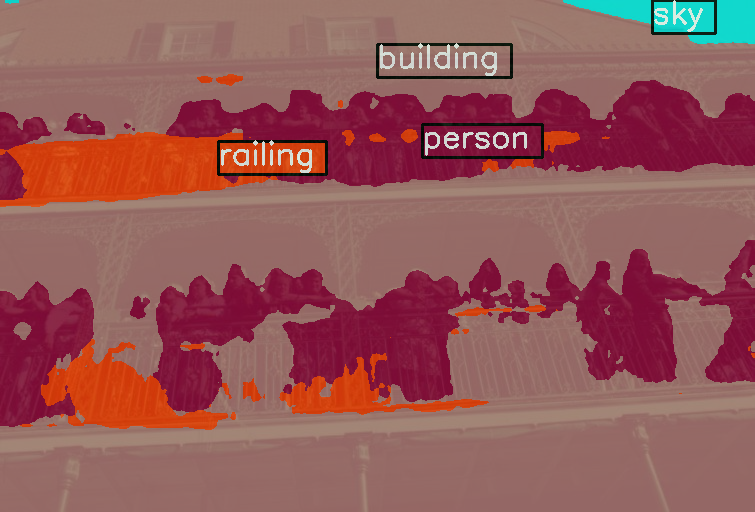}
        \end{subfigure}
         \begin{subfigure}{0.32\textwidth}
            \includegraphics[width=\linewidth]{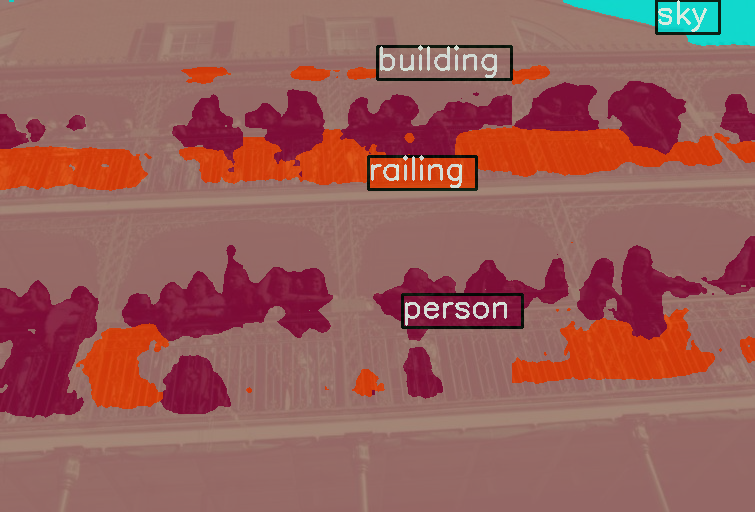}
        \end{subfigure}

        \begin{subfigure}{0.32\textwidth}
            \includegraphics[width=\linewidth]{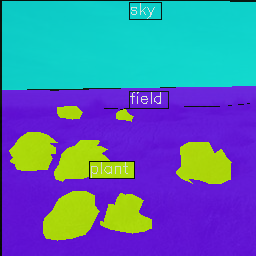}
            \caption{GT}
        \end{subfigure}
         \begin{subfigure}{0.32\textwidth}
            \includegraphics[width=\linewidth]{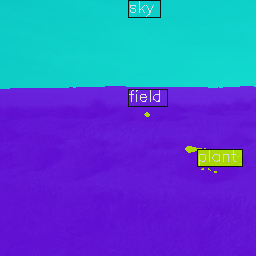}
            \caption{RAW-Adapter~\cite{raw_adapter}}
        \end{subfigure}
         \begin{subfigure}{0.32\textwidth}
            \includegraphics[width=\linewidth]{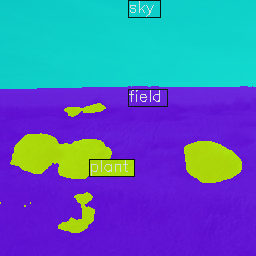}
            \caption{Our SimROD}
        \end{subfigure}
        
    \end{minipage}
    \caption{Semantic segmentation visualization results on ADE 20K RAW~\cite{raw_adapter}.}
    \label{fig:vis_sup_segm}
\end{figure*}

Comparison of SimROD and RAW-Adapter on the segmentation dataset are provided in Figure~\ref{fig:vis_sup_segm}. 

\subsection{More analysis of our GGE}
\label{sec_sup_gge}
\subsection{Gamma Transformation}
Gamma transformation is a nonlinear transformation to adjust the brightness and contrast of images. It is defined mathematically as:

\[
s = c \cdot r^\gamma
\]

where \( r \) represents the input pixel intensity normalized to \([0, 1]\), \( s \) is the transformed pixel intensity, \( c \) is a scaling constant (commonly \( c = 1 \)), and \( \gamma \) is the gamma value controlling the transformation.

For \( \gamma < 1 \), the transformation enhances dark regions (Gamma Compression), making the image brighter. Conversely, for \( \gamma > 1 \), it emphasizes bright regions (Gamma Expansion), resulting in a darker image. When \( \gamma = 1 \), no transformation is applied.

Our GGE module achieves learnable Gamma Transformation using only four parameters. With an almost negligible increase in parameter count and computational overhead, it achieves detection performance comparable to state-of-the-art (SOTA) methods.

\subsection{Hyperparameters Sensitivity Analysis}
The hyperparameters $\gamma_{min}$ and $\gamma_{max}$ are defined in our Global Gamma Enhancement (GGE). We have set their values by default according to ~\cite{xu2023toward}. To further analyze the GGE module, we have conducted hyperparameter sensitivity experiments on both LOD~\cite{hong2021crafting} and Pascal-Raw~\cite{omid2014pascalraw}. 

As shown in Figure~\ref{fig:gamma_lod} and Figure~\ref{fig:gamma_pas}, the results reveal that changes in \textit{gamma\_min} within the tested range ([0.08, 0.13]) have a negligible impact on mAP. Similarly, varying \textit{gamma\_max} across different values (0.1, 0.143, 0.25) results in minimal performance differences. This indicates that the model exhibits robust performance and low sensitivity to the choice of gamma range in this experimental setup. These findings suggest that the gamma transformation within the explored parameter bounds does not significantly alter the key features critical for object detection.

\subsection{Ablation Study}
The results presented in Table~\ref{tab:abla_rod} show that both proposed components—GGE and GGLE—exhibit consistent behavior across all evaluated input resolutions (640$\times$640, 960$\times$960, and 1280$\times$1280). Crucially, using GGLE alone leads to near-complete failure (e.g., mAP $\approx$ 0), indicating that local enhancement without global intensity normalization is ineffective. In contrast, GGE alone already provides substantial improvements over the baseline, and further combining it with GGLE consistently yields the best performance. This demonstrates that GGE is essential for object detection on the ROD~\cite{xu2023toward} dataset, while GGLE is only beneficial when built upon GGE. To better utilize GPU memory capacity, the experiments reported in this table were conducted using 4 GPUs with a total batch size of 16 (i.e., 4 per GPU), which slightly differs from the hyperparameters used in the main text. The necessity of GGE can be explained by the structure of ROD: daytime and nighttime images are interleaved in the same dataset, resulting in extreme pixel intensity discrepancies. Without a global adaptive normalization like GGE, the model cannot reconcile these divergent distributions, rendering subsequent local enhancements (e.g., GGLE) ineffective or even harmful.

\begin{table}[htbp]
  \centering
  \caption{\textbf{Ablation Study on ROD~\cite{xu2023toward}.} Bold numbers indicate the best result.}
    \begin{tabular}{ccc|cc}
    \toprule
    Input Size & GGE & GGLE & mAP & AP$_{50}$ \\
    \midrule
    \multirow{3}[2]{*}{1280$\times$1280} & \CIRCLE   &  \Circle   & 49.5 & 76.7 \\
        &  \Circle  &  \CIRCLE   & 0.0 & 0.0 \\
        &  \CIRCLE  &   \CIRCLE  & \textbf{51.5} & \textbf{78.9} \\
    \midrule
    \multirow{3}[2]{*}{960$\times$960} & \CIRCLE   &   \Circle  & 43.8 & 70.9 \\
        &  \Circle  &  \CIRCLE   & 0.0 & 0.2 \\
        & \CIRCLE   &  \CIRCLE   & \textbf{45.6} & \textbf{73.0} \\
    \midrule
    \multirow{3}[2]{*}{640$\times$640} & \CIRCLE   &  \Circle   & 32.4 & 56.3 \\
        &  \Circle  &  \CIRCLE   & 0.1 & 0.4 \\
        & \CIRCLE   &  \CIRCLE   & \textbf{34.8} & \textbf{59.5} \\
    \bottomrule
    \end{tabular}%
  \label{tab:abla_rod}%
\end{table}%

\subsection{Comparison between sRGB and RAW.}As shown in Table~\ref{tab:re_srgb_raw}, using sRGB after the ISP pipeline with GGE produces better results than RAW data with GGE. However, it still performs worse than our approach with GGLE. This result is expected, as the object detector is pretrained on sRGB data. It's also worth noting that using RAW data for object detection can eliminate the need for the ISP module, simplifying the pipeline and reducing hardware costs.
\begin{table}[htbp]
  \centering
  \caption{ Comparison between sRGB and RAW on LOD\cite{hong2021crafting} and Pascal-Raw~\cite{omid2014pascalraw}.}
    \begin{tabular}{ccccccc}
    \toprule
    \multirow{2}[4]{*}{Data} & \multirow{2}[4]{*}{GGE} & \multirow{2}[4]{*}{GGLE} & \multicolumn{2}{c}{LOD} & \multicolumn{2}{c}{Pascal-Raw} \\
\cmidrule{4-7}        &     &     & mAP & AP$_{50}$ & mAP & AP$_{50}$ \\
    \midrule
    sRGB &  \CIRCLE   &  \Circle   & 25.6 & 44.6 & \textbf{70.3} & 94.5 \\
    RAW &   \CIRCLE  &  \Circle    & 25.5 & 46.2 & 68.4 & 94.3 \\
    RAW &  \CIRCLE   &  \CIRCLE   & \textbf{26.7} & \textbf{46.3} & 69.7 & \textbf{95.1} \\
    \bottomrule
    \end{tabular}%
  \label{tab:re_srgb_raw}%
\end{table}%

\subsection{Latency}
\label{sec_latency}
As shown in Tab.~\ref{tab:inf_time}, we compare our method with DIAP~\cite{xu2023toward} and Raw-Adaptor~\cite{raw_adapter} on mAP on the LOD~\cite{hong2021crafting} dataset, parameter count, and latency. Our method achieves higher mAP while drastically reducing parameters and inference latency across different architectures. These results demonstrate the clear advantage of our approach in efficiency and performance.
\begin{table}[htbp]
  \centering
  \caption{Comparison of Latency and Parameters on the LOD.}
    \begin{tabular}{ccccc}
    \toprule
    Method & Detector & mAP & Params (M) & Latency \\
    \midrule
    DIAP & \multirow{2}[2]{*}{YOLOX} & 25.9 & 0.260  & 8.80ms \\
    Ours &     & 26.7 & 0.003  & 7.08ms \\
    \midrule
    Raw-Adapter & \multirow{2}[2]{*}{RetinaNet} & 62.1 & 0.460  & 0.960s \\
    Ours &     & 63.4 & 0.003  & 0.351s \\
    \bottomrule
    \end{tabular}%
  \label{tab:inf_time}%
\end{table}%

